\documentclass[fleqn,10pt]{wlscirep}
\usepackage[utf8]{inputenc}
\usepackage[T1]{fontenc}

\usepackage{amssymb}
\usepackage{amsmath}
\usepackage{array}
\usepackage{tabularx}
\usepackage{ragged2e}
\usepackage{enumitem}
\usepackage{rotating}
\usepackage{algorithm}
\usepackage{algpseudocode}
\usepackage{booktabs}
\usepackage{multirow}
\usepackage{xcolor}
\usepackage{textcomp}
\usepackage{manyfoot}
\usepackage[title]{appendix}
\usepackage{mathrsfs}
\usepackage{hyperref}
\usepackage{url}
\usepackage{doi}
\usepackage{amsthm}
\usepackage{listings}

\lstdefinelanguage{json}{
  basicstyle=\ttfamily\footnotesize,
  morestring=[b]",
  morestring=[d]',
  literate=
    *{0}{{{\color{red}0}}}{1}
     {1}{{{\color{red}1}}}{1}
     {2}{{{\color{red}2}}}{1}
     {3}{{{\color{red}3}}}{1}
     {4}{{{\color{red}4}}}{1}
     {5}{{{\color{red}5}}}{1}
     {6}{{{\color{red}6}}}{1}
     {7}{{{\color{red}7}}}{1}
     {8}{{{\color{red}8}}}{1}
     {9}{{{\color{red}9}}}{1}
     {:}{{{\color{blue}{:}}}}{1}
     {,}{{{\color{blue}{,}}}}{1}
     {\{}{{{\color{blue}{\{}}}}{1}
     {\}}{{{\color{blue}{\}}}}}{1}
     {[}{{{\color{blue}{[}}}}{1}
     {]}{{{\color{blue}{]}}}}{1},
}

\newcommand{\ProbNeural}{P_{\text{neural}}}
\newcommand{\ProbRule}{P_{\text{rule}}}
\newcommand{\ProbFinal}{P_{\text{final}}}

\theoremstyle{plain}

\theoremstyle{definition}

\theoremstyle{remark}

\title{EduRiskX: A Neuro-Symbolic Framework with F-Logic Reasoning for Early Academic Risk Prediction}

\author[1]{Yu Fu}
\author[1]{Yongqi Kang}
\author[1,*]{Yong Zhao}
\author[2,*]{Rongfang Bie}
\affil[1]{Sichuan University, College of Computer Science, Chengdu, 610207, China}
\affil[2]{Beijing Normal Univeristy, School of Artificial Intelligence, Beijing, 100875,China}


\begin{abstract}
Predicting students' academic risk in online education is crucial for enabling timely interventions that can improve retention and learning outcomes. However, existing models often suffer from limited early detection capability and insufficient interpretability, leading to a ``black-box'' trust crisis that hinders their adoption in real-world pedagogical settings. To address these challenges, we propose EduRiskX, a neuro-symbolic framework that integrates a temporal Transformer-based predictor with F-Logic symbolic reasoning. The neural component models longitudinal student activity sequences using temporal attention, class-weighted loss, and dynamic weekly truncation, and is trained independently on the 80\% training set. Acting as a data-driven modern expert system, an F-Logic rule base -- grounded in established educational theories (Engagement Theory and Student Integration Model) to mimic the diagnostic logic of human educators -- is constructed exclusively from the same training data. The neural risk probability and the symbolic confidence score are then combined through a logistic regression–based fusion mechanism that learns the relative contribution of each signal in a data-driven manner. Experiments on the Open University Learning Analytics Dataset (OULAD) using a strict 80/10/10 student-level split show that EduRiskX achieves an accuracy of 0.900 and an F1-score of 0.894 at the end of the semester (Week 38), with an average early detection week of 9.32 and a detection rate of 94.30\%. Compared with state-of-the-art time-series models (PatchTST, iTransformer) and commonly used deep learning baselines (LSTM, CNN), EduRiskX yields improved recall and earlier risk identification under identical conditions. Beyond predictive performance, the F-Logic module provides structured rule-based explanations that link risk predictions to observable behavioral patterns and educational theories, supporting transparent and pedagogically aligned early intervention. Our work demonstrates that neuro-symbolic integration offers a viable path toward trustworthy and actionable learning analytics systems.
\end{abstract}

\begin{document}
\flushbottom
\maketitle
\thispagestyle{empty}

\section*{Introduction}

The rapid expansion of online education has brought unprecedented opportunities for flexible learning, but also introduced significant challenges such as high dropout rates and heterogeneous student performance \cite{he2025lbtt}. Learning analytics aims to address these issues by leveraging behavioral data from digital platforms to identify students who may require additional academic support. Early and reliable identification of at-risk students is particularly critical, as timely interventions have been shown to mitigate learning difficulties and improve retention outcomes. However, two fundamental obstacles persist in current predictive systems.

First, many existing deep learning models—including Long Short-Term Memory (LSTM) networks, convolutional neural networks (CNNs), and recent Transformer-based architectures—rely on extensive accumulation of sequential data to achieve satisfactory performance \cite{xing2019dropout, alsariera2022assessment}. This dependency constrains their effectiveness for early-stage risk detection, often delaying identification until well into the semester when intervention windows have narrowed. Second, the limited interpretability of complex neural models reduces transparency and hampers educators' ability to connect predictions with actionable pedagogical strategies. Even state-of-the-art time-series models such as PatchTST \cite{nie2023timeseriesworth64} and iTransformer \cite{liu2024itransformerinvertedtransformerseffective}, while powerful for forecasting, are not designed to provide explanations that align with educational theory.

In recent years, neuro-symbolic artificial intelligence—which integrates neural networks with symbolic reasoning—has emerged as a promising direction for improving interpretability while maintaining predictive capability \cite{garcez2023neurosymbolic, rachha2023explainable}. By combining data-driven representation learning with structured logical inference, neuro-symbolic approaches offer a pathway toward balancing performance and explainability in educational analytics. Frame Logic (F-Logic) is particularly suitable for symbolic reasoning in this context due to its ability to represent object attributes, hierarchical structures, and rule-based inference in a declarative, executable manner \cite{10.1145/67544.66939}.

Building on this perspective, we propose EduRiskX---a neuro-symbolic framework that functions as a modern educational expert system by integrating an optimized temporal Transformer with an F-Logic symbolic reasoning module. By doing so, EduRiskX directly addresses the ``black-box'' trust crisis prevalent in pure deep learning models \cite{cai2025practices}, explicitly mimicking the diagnostic logic of human experts to bridge the gap between predictive accuracy and pedagogical applicability. The Transformer-based neural predictor is first trained to estimate student-level risk probabilities from longitudinal behavioral sequences. Concurrently, an F-Logic rule base is constructed exclusively from the 80\% training set, ensuring that no symbolic knowledge is derived from validation or test data. Finally, a logistic regression–based fusion mechanism adaptively combines neural risk probabilities and symbolic confidence scores, allowing their relative contributions to be learned in a data-driven manner.

The main contributions of this paper are as follows:

\begin{enumerate}[label=(\arabic*), leftmargin=*]
    \item \textbf{An optimized temporal Transformer architecture} incorporating temporal attention, class-weighted loss, and dynamic weekly truncation to improve modeling of longitudinal behavioral sequences and address class imbalance, particularly in early prediction stages.

    \item \textbf{A training-set–grounded F-Logic reasoning module} that formalizes pedagogically informed risk patterns as symbolic rules and produces interpretable confidence scores. The rule base is derived from educational theories (Engagement Theory \cite{fredricks2004school} and Student Integration Model \cite{tinto1975dropout}) and is constructed using a systematic rule-mining procedure that avoids data leakage.

    \item \textbf{A learnable logistic fusion strategy} that integrates neural predictions and symbolic reasoning outputs, enabling the model to calibrate the contribution of each evidence source and improve early detection performance while preserving interpretability.

    \item \textbf{Structured explanation outputs for pedagogical decision support}, including triggered rules, theory-aligned behavioral interpretations, and mapped intervention suggestions, facilitating transparent and actionable early intervention in real-world educational settings.
\end{enumerate}

The remainder of this paper is organized as follows. Section~\ref{sec:related_work} reviews related work on academic risk prediction and neuro-symbolic learning. Section~\ref{sec:framework} presents the EduRiskX framework in detail. Section~\ref{sec:results} describes the experimental setup and evaluation results. Section~\ref{sec:case_study} provides interpretability case analyses. Section~\ref{sec:discussion} discusses implications and limitations. Finally, Section~\ref{sec:conclusion} concludes the paper and outlines future work.

\section*{Related Work}
\label{sec:related_work}

This section situates our work within three interrelated research streams: (1) predictive modeling for student academic risk using the Open University Learning Analytics Dataset (OULAD), particularly sequential deep learning approaches; (2) neuro-symbolic artificial intelligence (AI) in educational contexts; and (3) F-Logic-based symbolic reasoning for interpretable AI systems.

\subsection*{Predictive Modeling with the OULAD Dataset}
\label{subsec:related_oulad}

The Open University Learning Analytics Dataset (OULAD) has become a widely used benchmark in educational data mining due to its scale, longitudinal structure, and realistic representation of distance higher education \cite{kuzilek2017open}. Early studies on OULAD primarily employed traditional machine learning methods, such as logistic regression and decision trees, based on aggregated behavioral indicators \cite{romero2010educational, berland2014educational}. These approaches established baseline performance and demonstrated the feasibility of data-driven academic risk prediction.

As interest in temporal modeling increased, researchers began applying deep learning techniques to capture sequential learning behaviors \cite{hernandez2019systematic}. Recurrent Neural Networks (RNNs) and Long Short-Term Memory (LSTM) models have been widely adopted to model weekly or event-level clickstream sequences, generally reporting improved predictive performance compared with static classifiers \cite{WAHEED2023118868}. More recently, Transformer-based architectures have achieved state-of-the-art (SOTA) results by leveraging self-attention mechanisms to model long-range temporal dependencies \cite{10047866}. Hybrid deep learning frameworks further enhance performance through attention mechanisms or multi-feature fusion strategies. For example, the LBTT model \cite{he2025lbtt} integrates time-window representations and behavior-type information via a dual-attention mechanism.

Despite these advances, two challenges remain under active investigation. First, while deep neural models provide strong predictive capability, their internal representations are often difficult to interpret directly in pedagogical terms. Second, performance in early-course stages may vary depending on the amount of available sequential data. These considerations highlight the importance of developing approaches that maintain competitive predictive performance while improving interpretability and early-stage robustness.

\subsection*{Neuro-Symbolic AI in Education}
\label{subsec:related_xai}

Improving transparency in educational predictive models has led to increasing interest in neuro-symbolic artificial intelligence (AI), which combines neural networks with symbolic reasoning \cite{garcez2023neurosymbolic}. By integrating data-driven learning with structured logical inference, neuro-symbolic approaches aim to balance predictive accuracy and interpretability \cite{rachha2023explainable}.

\textbf{Inherently interpretable models.} Rule-based systems represent one of the earliest forms of interpretable AI in education. In intelligent tutoring systems, domain knowledge has traditionally been encoded as explicit rules, enabling direct inspection of decision logic \cite{nwana1990intelligent}. In educational data mining, data-driven rule discovery techniques such as Association Rule Mining (ARM) have been applied to identify frequent behavioral patterns \cite{agrawal1993mining}. However, when applied to high-dimensional and sequential datasets, rule mining may require careful constraint design to ensure pedagogical relevance and manageable rule complexity.

\textbf{Post-hoc explanation methods.} Post-hoc explanation techniques, including LIME and SHAP, have been widely used to provide feature attribution for black-box models \cite{10.1145/2939672.2939778}. These methods offer local explanations by estimating the contribution of input variables to individual predictions. While useful for interpretability and model analysis, post-hoc explanations approximate model behavior and may not directly encode structured domain knowledge.

\textbf{Neural-symbolic integration.} More recent work explores tighter integration between neural representation learning and symbolic reasoning. Such approaches vary in architectural design and integration strategy, and ongoing research examines trade-offs among scalability, predictive performance, and interpretability. Emerging techniques, including rule-informed prompt tuning and logic-guided learning \cite{chen2022knowprompt}, illustrate alternative mechanisms for incorporating structured knowledge into neural systems, although applications in academic risk prediction remain limited. EduRiskX follows a modular integration strategy in which neural prediction and symbolic reasoning are trained separately and subsequently combined, allowing each component to retain its strengths while enabling interpretable output.

\subsection*{F-Logic for Symbolic Reasoning in Education}
\label{subsec:related_theory}

Frame Logic (F-Logic) provides a declarative formalism for representing objects, attributes, and inference rules in a structured and executable manner \cite{10.1145/66926.66939, 10.1145/67544.66939}. Its support for object-oriented representation and rule chaining makes it suitable for modeling educational constructs such as engagement patterns, participation dynamics, and academic integration.

Although symbolic logic systems have been applied in educational knowledge representation, their integration with modern deep neural predictors for academic risk prediction remains relatively underexplored. Existing studies often focus either on purely neural approaches or on standalone rule-based systems. EduRiskX contributes to this line of research by integrating an F-Logic-based symbolic reasoning module—constructed exclusively from the training set—with a temporal Transformer predictor, using a logistic regression–based fusion mechanism to combine complementary predictive signals. This design aims to maintain competitive predictive performance while enhancing interpretability and pedagogical transparency.

\section*{Methods}
\label{sec:framework}

\begin{figure}[!t]
\centering
\includegraphics[width=\linewidth]{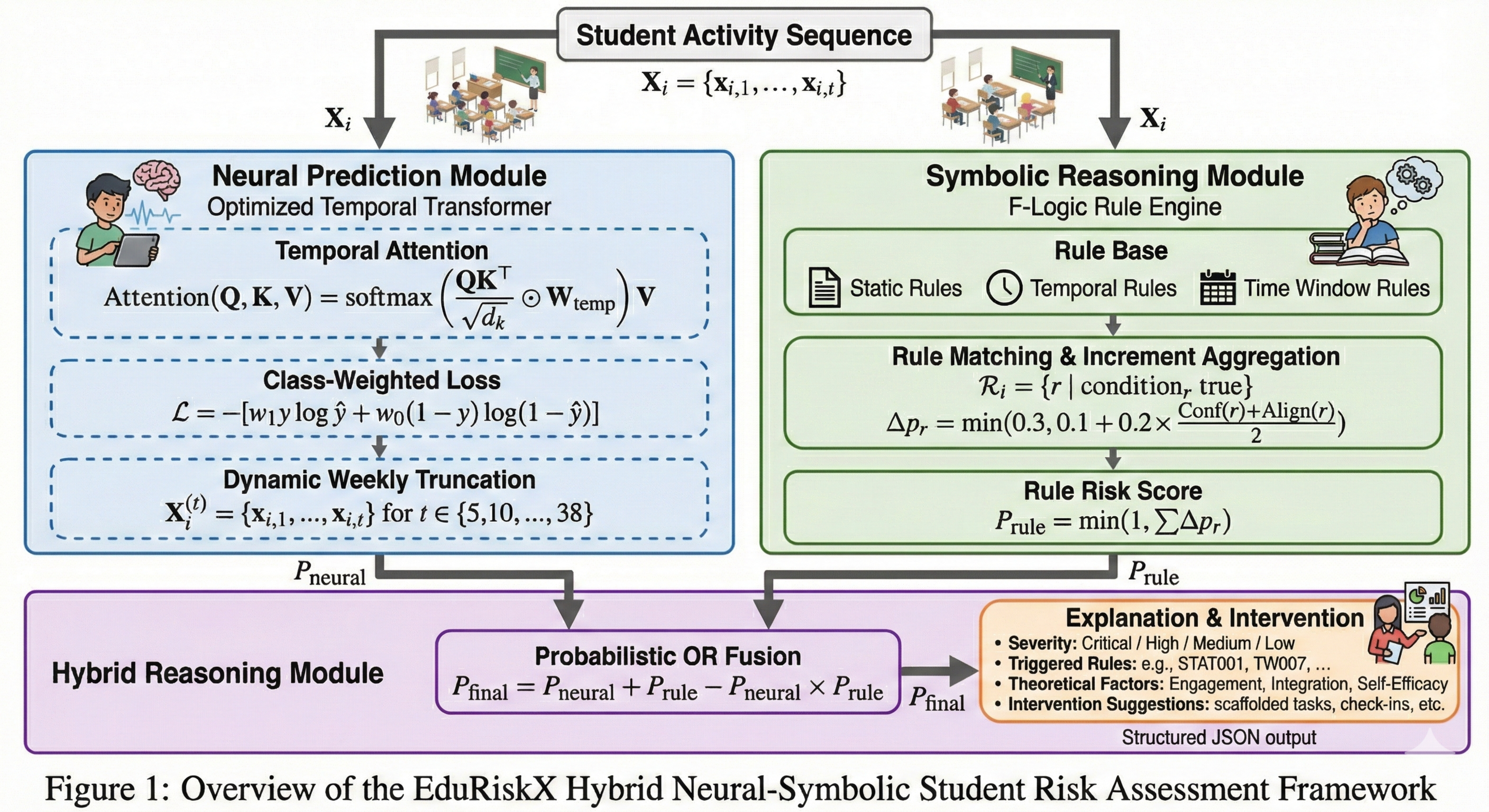}
\caption{Overall architecture of the EduRiskX framework. \textbf{Left (Neural Prediction Module)}: Optimized Temporal Transformer incorporating Temporal Attention, Class-Weighted Loss, and Dynamic Weekly Truncation to produce $\ProbNeural$. \textbf{Right (Symbolic Reasoning Module)}: F-Logic rule engine that applies 120 pedagogically-grounded rules (Static, Temporal, Time Window) to compute $\ProbRule$ via rule confidence and risk increment calibration. \textbf{Bottom (Hybrid Inference)}: Two-stage logistic fusion $\ProbFinal = \sigma(\alpha \cdot \text{logit}(\ProbNeural) + \beta \cdot \text{logit}(\ProbRule) + b)$, followed by explanation generation (risk severity, triggered rules, theoretical alignment, intervention suggestions).}
\label{fig:framework_overview}
\end{figure}

EduRiskX adopts a two-stage ``Predict-then-Explain'' neuro-symbolic architecture, comprising two core components: a neural prediction module (optimized temporal Transformer) and a symbolic reasoning module (F-Logic). The overall framework is illustrated in Figure~\ref{fig:framework_overview}.

\subsection*{Data Preprocessing and Temporal Feature Construction}
\label{subsec:framework_data}

EduRiskX operates on longitudinal student interaction data collected from the Open University Learning Analytics Dataset (OULAD) \cite{kuzilek2017open}. Raw event-level logs are first aggregated into fixed-length weekly sequences to align with the pedagogical structure of academic courses. For each student, we construct a multivariate time series:
\[
\mathbf{X}_i = \{\mathbf{x}_{i,1}, \mathbf{x}_{i,2}, \dots, \mathbf{x}_{i,T}\},
\]
where $\mathbf{x}_{i,t} \in \mathbb{R}^d$ represents a $d$-dimensional vector of behavioral and performance features for student $i$ during week $t$, and $T$ denotes the total number of instructional weeks (38 weeks for OULAD). Features include measures of content access, assessment activity, forum participation, and engagement intensity. Missing weeks are explicitly encoded to preserve temporal continuity and enable early-stage prediction.

This weekly temporal representation serves as a unified input format for all downstream predictive models, enabling fair comparison across baselines and advanced architectures.

\subsection*{Neural Prediction Module (Optimized Transformer)}
\label{subsec:framework_predictor}

The neural prediction module is an optimized Transformer-based neural network, specifically designed for the temporal sequence characteristics of student online behavior data. Building on the standard Transformer encoder, we introduce three key optimizations to enhance prediction performance and training stability.

\begin{enumerate}[label=(\arabic*), leftmargin=*]
    \item \textbf{Temporal Attention}: The standard self-attention mechanism is augmented with a learnable temporal weight matrix $\mathbf{W}_{\text{temp}} \in \mathbb{R}^{T \times T}$. For each head, the attention scores are computed as:
    \begin{equation}
       \mathrm{Attention}(\mathbf{Q},\mathbf{K},\mathbf{V}) = \mathrm{softmax}\left(\frac{\mathbf{Q}\mathbf{K}^\top}{\sqrt{d_k}} \odot \mathbf{W}_{\text{temp}}\right)\mathbf{V} 
    \end{equation}
    where $\odot$ denotes element-wise multiplication. $\mathbf{W}_{\text{temp}}$ is jointly trained with the model and assigns higher weights to pedagogically critical weeks (e.g., assessment deadlines) and lower weights to distant past weeks, enabling explicit modeling of long-range temporal dependencies.

    \item \textbf{Class-Weighted Loss}: To address the inherent class imbalance (at-risk students constitute only $\sim$12\% of the population), we employ a weighted cross-entropy loss:
    \begin{equation}
        \mathcal{L} = -\frac{1}{N}\sum_{i=1}^{N} \left[ w_1 \cdot y_i \log\hat{y}_i + w_0 \cdot (1-y_i)\log(1-\hat{y}_i) \right]
    \end{equation}
    where $w_1 = \frac{\#\text{non-risk}}{\#\text{risk}}$ and $w_0 = 1$. This weighting imposes a higher penalty on false negatives, thereby improving recall—the most critical metric for early warning systems.

    \item \textbf{Dynamic Weekly Truncation}: To simulate realistic early prediction scenarios, we train and fine-tune separate model instances for each prediction week $t \in \{5,10,15,20,25,30,35,38\}$. During training for a specific cutoff week $t$, the input sequence is truncated to the first $t$ weeks: $\mathbf{X}_i^{(t)} = \{\mathbf{x}_{i,1},\dots,\mathbf{x}_{i,t}\}$. This ensures that the model learns to make predictions based on the exact information available at that point in the semester, without any future data leakage.
\end{enumerate}

\noindent
\textbf{Computation of $\ProbNeural$.} Given a truncated sequence $\mathbf{X}_i^{(t)} \in \mathbb{R}^{t \times d}$, the following steps produce the neural risk probability:
\begin{enumerate}[label=\textit{Step \arabic*:}, leftmargin=*]
    \item \textit{Embedding}: A linear projection maps the input to $d_{\text{model}}$-dimensional embeddings: $\mathbf{E} = \mathbf{X}_i^{(t)}\mathbf{W}_e + \mathbf{b}_e$, where $\mathbf{W}_e \in \mathbb{R}^{d \times d_{\text{model}}}$.
    \item \textit{Positional Encoding}: Learnable positional encodings $\mathbf{P} \in \mathbb{R}^{t \times d_{\text{model}}}$ are added: $\mathbf{Z}^{(0)} = \mathbf{E} + \mathbf{P}$.
    \item \textit{Transformer Encoder}: $L$ encoder layers (each consisting of multi-head temporal attention and feed-forward network with residual connections) produce contextualized representations $\mathbf{Z}^{(L)} \in \mathbb{R}^{t \times d_{\text{model}}}$.
    \item \textit{Output Projection}: The representation at the \textit{last} time step is extracted and passed through a linear layer followed by sigmoid activation:
    \begin{equation}
        \ProbNeural = \sigma\left( \mathbf{z}^{(L)}_t \mathbf{W}_o + b_o \right)
    \end{equation}
    where $\mathbf{z}^{(L)}_t$ is the $t$-th row of $\mathbf{Z}^{(L)}$, $\mathbf{W}_o \in \mathbb{R}^{d_{\text{model}} \times 1}$, $b_o \in \mathbb{R}$, and $\sigma(\cdot)$ is the logistic sigmoid function.
\end{enumerate}
The entire module $f_\theta$ (with parameters $\theta = \{\mathbf{W}_e,\mathbf{b}_e,\mathbf{P},\text{encoder},\mathbf{W}_o,b_o\}$) is trained end-to-end using the class-weighted loss.

\subsection*{Symbolic Reasoning Module (F-Logic)}
\label{subsec:framework_reasoning}

The symbolic reasoning module enhances the interpretability and early detection capability of the framework by mining and inferring academic risk rules based on F-Logic. It comprises two core functions: rule mining and hybrid inference.

\subsubsection*{Rule Mining Procedure}
\label{subsubsec:rule_mining}

We employ a combinatorial search algorithm inspired by the Apriori principle, but optimized for feature constraints and educational interpretability. The algorithm iteratively generates rule candidates of length \(k = 1\) to \(4\) using \texttt{itertools.combinations} and evaluates them based on support and confidence. All mining procedures are conducted exclusively on the 80\% training set to prevent information leakage. The key parameters are:
\begin{itemize}[leftmargin=*]
    \item Minimum support: \(0.01\) (approximately 30 students) to ensure statistical reliability.
    \item Minimum confidence: \(0.35\) for static co-occurrence rules, \(0.65\) for temporal trend rules (reflecting the higher reliability of sequential patterns), and \(0.25\) for time-window rules (to capture early subtle signals).
    \item Maximum rule length: 4 features.
\end{itemize}

The general form of a rule is:
\[
\mathrm{IF}\; \mathrm{Condition}_1 \land \mathrm{Condition}_2 \land \dots \land \mathrm{Condition}_n \;\mathrm{THEN}\; \text{RiskProb} \mathrel{+}= \Delta p
\]
where $\mathrm{Condition}_i$ represents a student behavior condition (e.g., $\text{submission\_delay} > 2$, $\text{quiz\_score} < 40$), and $\Delta p$ is the risk probability increment triggered by the rule. After mining, we apply a redundancy removal step: rules are sorted by descending confidence, and any rule whose feature set is a subset of a higher-confidence rule is discarded. This ensures that the rule base contains only the most concise and predictive patterns. The final rule base consists of 120 pedagogically grounded rules, with each rule assigned a confidence score based on its empirical validity in the training data (Equation~\ref{eq:rule_confidence}). 

To strictly prevent data leakage, all rule mining procedures—including support computation, confidence estimation, and threshold selection—are conducted exclusively on the 80\% training subset defined in the 80/10/10 train–validation–test split. Validation and test sets are never accessed during rule discovery. Once mined, the final rule base of 120 rules is frozen and applied unchanged to validation and test data.

\subsubsection*{Hybrid Inference with Two-Stage Logistic Fusion}
\label{subsubsec:fusion}

To avoid the conditional independence assumption implied by probabilistic OR and to enable adaptive weighting between neural and symbolic evidence, EduRiskX adopts a learnable logistic fusion mechanism implemented in a two-stage training framework.

\textbf{Stage 1: Neural Pretraining.}
The optimized Transformer module $f_\theta$ is first trained independently using the class-weighted loss defined in Section~\ref{subsec:framework_predictor}. After convergence, all neural parameters are frozen.

\textbf{Stage 2: Fusion Learning.}
Given frozen neural predictions $\ProbNeural$ and rule-based probabilities $\ProbRule$, the final risk probability is computed in logit space:

\begin{equation}
\label{eq:logit_def}
\mathrm{logit}(p) = \log \frac{p}{1-p}
\end{equation}

\begin{equation}
\label{eq:logistic_fusion}
\ProbFinal =
\sigma\big(
\alpha \cdot \mathrm{logit}(\ProbNeural)
+
\beta \cdot \mathrm{logit}(\ProbRule)
+
b
\big)
\end{equation}

where $\alpha$, $\beta$, and $b$ are learnable scalar parameters.

Fusion parameters are optimized using binary cross-entropy:

\begin{equation}
\label{eq:fusion_loss}
\mathcal{L}_{\text{fusion}}
=
-\frac{1}{N}
\sum_{i=1}^{N}
\left[
y_i \log \ProbFinal
+
(1-y_i)\log(1-\ProbFinal)
\right].
\end{equation}

During Stage 2, neural encoder parameters and rule definitions remain fixed, ensuring modularity and preventing representation contamination.

This design provides three advantages:

\begin{enumerate}[label=(\arabic*), leftmargin=*]
    \item It removes the independence assumption of probabilistic OR, allowing the model to account for correlations between neural and rule-based signals.
    \item It enables adaptive weighting: the learned coefficients $\alpha$ and $\beta$ reflect the relative reliability of each evidence source.
    \item It improves probability calibration through logit-space scaling, yielding well-calibrated final risk estimates.
\end{enumerate}

Thus, the hybrid module functions as a learnable evidence combiner rather than a fixed logical operator.

\subsubsection*{Rule Confidence and Risk Increment Calibration}
\label{subsubsec:risk_increment}

Each mined rule $r$ is assigned a statistical confidence score based on its empirical validity in the training data:
\begin{equation}
\label{eq:rule_confidence}
\mathrm{Conf}(r) = \frac{\#(\text{conditions}_r \land \text{at-risk})}{\#(\text{conditions}_r)},
\end{equation}
where $\#(\text{conditions}_r)$ is the number of students satisfying the rule antecedent, and $\#(\text{conditions}_r \land \text{at-risk})$ is the count of such students who are ultimately labeled as at-risk. 

To balance statistical reliability and pedagogical coverage, we adopt a two-tier selection strategy. 

\begin{itemize}[leftmargin=*]
    \item \textbf{Primary rules} satisfy $\mathrm{Support}(r) > 0.1$ and $\mathrm{Conf}(r) \geq 0.7$.
    \item \textbf{Auxiliary rules} satisfy $0.3 \leq \mathrm{Conf}(r) < 0.7$ and are retained when strongly aligned with established educational theories (theory alignment score $\ge 0.2$, see Section~\ref{subsec:framework_rules}).
\end{itemize}

This strategy preserves high-confidence risk indicators while maintaining early-term sensitivity.

To convert a triggered rule into a quantifiable contribution to the rule-based risk score, we map its confidence and theory alignment score to a risk increment $\Delta p_r$ via a calibrated linear function:
\begin{equation}
\label{eq:risk_increment}
\Delta p_r = \min\left(0.3,\; 0.1 + 0.2 \times \frac{\mathrm{Conf}(r) + \mathrm{Align}(r)}{2}\right),
\end{equation}
where $\mathrm{Align}(r) \in [0,1]$ is the theory alignment score (Section~\ref{subsec:framework_rules}). 
This mapping bounds $\Delta p_r$ between $0.1$ and $0.3$ per rule. The coefficients (0.1, 0.2, 0.3) were heuristically chosen to ensure that $\Delta p_r$ remains within a reasonable range and were subsequently verified on the validation set to yield well-calibrated $\ProbRule$ scores that are comparable in scale to $\ProbNeural$. 

The total rule-based risk score is then the sum of increments from all triggered rules, capped at $1.0$:
\begin{equation}
\label{eq:rule_score}
\ProbRule = \min\left(1.0,\; \sum_{r \in \mathcal{R}_i} \Delta p_r\right).
\end{equation}

\subsection*{F-Logic Rule Base for Academic Risk Assessment}
\label{subsec:framework_rules}

We developed a comprehensive F-Logic rule base consisting of 120 pedagogically-grounded rules to identify students at risk of poor academic performance. Among these, 16 rules are designated as \textit{strongly aligned} (theory alignment score $\ge 0.3$), indicating particularly close correspondence with core theoretical constructs. The rules are grounded in \textit{Engagement Theory} \cite{fredricks2004school} (behavioral persistence in academic activities) and the \textit{Student Integration Model} \cite{tinto1975dropout} (social-academic integration). The rule base comprises three categories: Static Pattern Rules, Temporal Sequence Rules, and Time Window Rules, each serving distinct diagnostic purposes.

\textbf{Theory Alignment Score Quantification:} To objectively measure how well a rule reflects educational theory, we compute an automated alignment score using semantic similarity. For a rule with feature set $F$, we define:
\[
\mathrm{Align}(r) = 0.4 \times S_{\text{keyword}} + 0.6 \times S_{\text{semantic}},
\]
where $S_{\text{keyword}}$ is the proportion of theoretical keywords (e.g., ``engagement'', ``integration'', ``self-efficacy'') that appear in the rule's feature names, and $S_{\text{semantic}}$ is the cosine similarity between the SBERT embedding \cite{reimers2019sentence} of the rule's feature description and the embedding of the corresponding theory's core constructs. We use the pre-trained \texttt{all-MiniLM-L6-v2} model for generating embeddings. This automated approach ensures consistency and avoids subjectivity in expert scoring.

\begin{figure}[t]
    \centering
    \includegraphics[width=\textwidth]{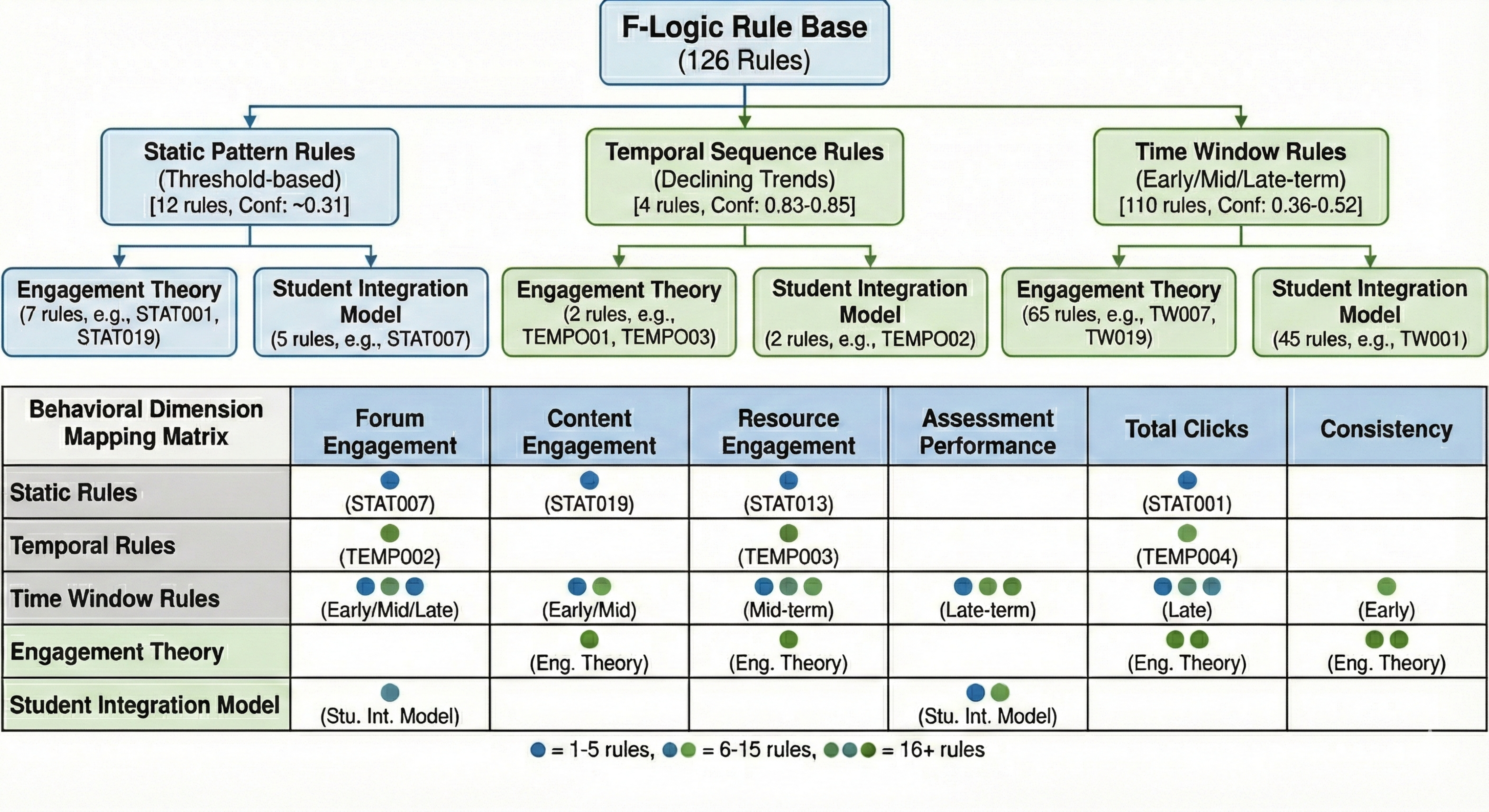}
    \caption{Overview of the F-Logic rule base used in EduRiskX. The matrix shows the mapping between rule types (Static, Temporal, Time Window), behavioral dimensions, and educational theories. Circle sizes indicate the number of rules in each category, and color intensity (blue) represents the average theoretical alignment score (darker = higher alignment). The rightmost column shows the average alignment score per rule category. E = Engagement Theory, S = Student Integration Model.}
    \label{fig:rule_base_overview}
\end{figure}

Figure~\ref{fig:rule_base_overview} provides a systematic visualization of the structure and coverage of the pedagogically-grounded rule base. The matrix-based layout elucidates the mapping between three core rule types (Static, Temporal, and Time Window), key behavioral dimensions in online learning (e.g., forum participation, content access, assessment performance), and their underlying educational theoretical constructs (Engagement Theory and Student Integration Model).

The horizontal axis represents behavioral dimensions, while the vertical axis categorizes rules by their temporal logic and detection mechanism. The shading intensity (or marker size) within each cell indicates the relative number of rules addressing that specific behavior-rule type combination. A separate column summarizes the average theoretical alignment score for each rule category, quantitatively reflecting the extent to which rule design is grounded in established educational theory.

This visualization highlights several key design principles of the rule base: 
(1) \textbf{Temporal Coverage}: Time Window rules provide dense coverage across all behavioral dimensions, enabling phase-specific risk detection; 
(2) \textbf{Theoretical Foundation}: Rules are explicitly mapped to theoretical constructs, ensuring that risk signals are pedagogically interpretable; 
(3) \textbf{Behavioral Specificity}: Different rule types target distinct behavioral patterns, from persistent low engagement (Static) to declining trends (Temporal). 

\subsection*{Behavior-Rule Mapping Framework}

Table~\ref{tab:behavior-rule-simple} presents a pedagogical mapping between key behavioral dimensions observed in online learning environments and the corresponding F-Logic rule categories implemented in EduRiskX. This mapping aids educators in understanding how specific student behaviors translate into interpretable risk signals.

\begin{table}[h]
\centering
\caption{EduRiskX: Representative Behavior-Rule Mappings}
\label{tab:behavior-rule-simple}
\footnotesize
\begin{tabular}{|l|l|l|r|r|}
\hline
\textbf{Behavior} & \textbf{Rule ID} & \textbf{Type} & \textbf{Conf.} & \textbf{Sup.} \\
\hline
Total Clicks (VL) & STAT001 & Static & 0.312 & 0.611 \\
\hline
Forum Decline & TEMP002 & Temporal & 0.850 & 0.779 \\
\hline
Resource Decline & TEMP003 & Temporal & 0.833 & 0.821 \\
\hline
Early Low Content & TW007 & Time Window & 0.438 & 1.000 \\
\hline
Low Consistency & TW019 & Time Window & 0.468 & 0.342 \\
\hline
Quiz $< 40$ & TW079 & Time Window & 0.346 & 0.934 \\
\hline
Delay $> 2$ wks & TW108 & Time Window & 0.346 & 0.934 \\
\hline
No Forum $\times 4$ & STAT007 & Static & 0.335 & 0.423 \\
\hline
3-week Trend $< 0$ & TEMP001 & Temporal & 0.850 & 0.907 \\
\hline
Early Low Forum & TW001 & Time Window & 0.520 & 0.300 \\
\hline
\end{tabular}
\vspace{2pt}
\footnotesize
\textit{Note}: Conf. = Confidence, Sup. = Support. Based on 120 F-Logic rules.
\end{table}

\textbf{Static Pattern Rules} rely on single-feature threshold values to identify persistent low engagement, serving as broad screening mechanisms during initial risk assessment. The foundational static rule (STAT001) targets students with $\text{total\_clicks\_cat} = \text{'VL'}$ (Very Low total clicks), covering 331,923 students (support = 0.611) with confidence 0.312. Additional static rules identify students with zero forum participation (STAT007), zero resource access (STAT013), and zero content engagement (STAT019), each aligned with specific theoretical constructs.

\textbf{Temporal Sequence Rules} detect declining trends in engagement metrics over consecutive time periods, representing the most reliable risk signals (confidence $\ge$ 0.833). Four core temporal rules were identified:
\begin{itemize}[leftmargin=*]
    \item \textbf{TEMP001}: Negative 3-week activity trend ($\text{activity\_trend\_3w} < 0$ for $\ge$1 consecutive periods), confidence 0.85, support 0.907 (23,661 students), aligned with Engagement Theory \cite{fredricks2004school}.
    \item \textbf{TEMP002}: Declining forum clicks over consecutive periods, confidence 0.85, support 0.779 (20,310 students), aligned with Student Integration Model \cite{tinto1975dropout}.
    \item \textbf{TEMP003}: Declining resource clicks over consecutive periods, confidence 0.833, support 0.821 (21,415 students), aligned with Engagement Theory \cite{fredricks2004school}.
    \item \textbf{TEMP004}: Declining total clicks over consecutive periods, confidence 0.85, support 0.864 (22,524 students), aligned with Engagement Theory \cite{fredricks2004school}.
\end{itemize}

\textbf{Time Window Rules} identify risk signals within specific temporal windows, enabling early-term detection. The rule base includes rules for three time periods: early-term (Weeks 0-4), mid-term (Weeks 5-12), and late-term (Weeks 13-20), each targeting different engagement dimensions:

\textbf{Early-Term Rules (Weeks 0-4):} Critical for initial risk screening, these rules identify students exhibiting low engagement during the formative phase of the course. Key rules include:
\begin{itemize}[leftmargin=*]
    \item \textbf{TW001}: Low forum clicks ($\text{week} \in [0,4] \land \text{clicks\_forum} \le 0.0$), confidence 0.52, support 0.3 (32,041 students), aligned with Student Integration \cite{tinto1975dropout}.
    \item \textbf{TW007}: Low content clicks ($\text{week} \in [0,4] \land \text{clicks\_content} \le 0.0$), confidence 0.438, support 1.0 (106,760 students), aligned with Engagement \cite{fredricks2004school}.
    \item \textbf{TW019}: Low activity consistency ($\text{week} \in [0,4] \land \text{activity\_consistency} \le 1.0$), confidence 0.468, support 0.342 (36,490 students), highest theory alignment score (0.363), aligned with Engagement \cite{fredricks2004school}.
\end{itemize}

\textbf{Mid-Term Rules (Weeks 5-12):} These rules detect risk patterns during the consolidation phase of learning, including declining forum activity (TW030-031), low resource engagement (TW034-035), and inconsistent participation patterns (TW048-049).

\textbf{Late-Term Rules (Weeks 13-20):} Focus on identifying persistent risk factors that may lead to course failure, including sustained low engagement (TW063-072), poor assessment performance (TW078-082), and declining social integration (TW085-086).

\textbf{Entire-Course Rules (Weeks 0-20):} Cross-temporal rules that identify students exhibiting consistent risk patterns throughout the semester, including persistent low forum participation (TW087-088), sustained low content engagement (TW093-094), and chronic submission delays (TW107-109). These rules reflect prolonged deficits in self-regulated learning, which is closely associated with academic self-efficacy \cite{bandura1997self, locke1997self}.

Table~\ref{tab:detailed_risk_rules} provides a comprehensive overview of representative rules from the 120-rule base used in EduRiskX.

\begin{sidewaystable}[!htbp]
  \centering
  \caption{Representative Academic Risk Assessment Rules in EduRiskX}
  \label{tab:detailed_risk_rules}
  \footnotesize
  \begin{tabular}{@{}lcccclcccc@{}}
    \toprule
    \textbf{Rule ID} & \textbf{Type} & \textbf{Time Window} & \textbf{Condition} & \textbf{Confidence} & \textbf{Support} & \textbf{Affected} & \textbf{Theory} & \textbf{Align. Score} \\
    \midrule
    STAT001 & Static & Full & $\text{total\_clicks\_cat} = \text{'VL'}$ & 0.312 & 0.611 & 331,923 & Engagement & 0.146 \\
    STAT007 & Static & Full & $\text{clicks\_forum} = 0$ & 0.335 & 0.423 & 229,676 & Stud. Integ. & 0.209 \\
    \midrule
    TEMP001 & Temporal & Full & $\text{activity\_trend\_3w} < 0$ ($\ge$1) & 0.850 & 0.907 & 23,661 & Engagement & 0.224 \\
    TEMP002 & Temporal & Full & $\text{clicks\_forum}\downarrow$ consecutive & 0.850 & 0.779 & 20,310 & Stud. Integ. & 0.209 \\
    \midrule
    TW001 & Time & Weeks 0-4 & $\text{clicks\_forum} \le 0.0$ & 0.520 & 0.300 & 32,041 & Stud. Integ. & 0.209 \\
    TW007 & Time & Weeks 0-4 & $\text{clicks\_content} \le 0.0$ & 0.438 & 1.000 & 106,760 & Engagement & 0.100 \\
    TW019 & Time & Weeks 0-4 & $\text{activity\_consistency} \le 1.0$ & 0.468 & 0.342 & 36,490 & Engagement & 0.363 \\
    \midrule
    TW030 & Time & Weeks 5-12 & $\text{clicks\_forum} \le 0.0$ & 0.412 & 0.411 & 53,953 & Stud. Integ. & 0.209 \\
    TW048 & Time & Weeks 5-12 & $\text{activity\_consistency} \le 0.25$ & 0.345 & 0.277 & 36,437 & Engagement & 0.363 \\
    \midrule
    TW063 & Time & Weeks 13-20 & $\text{clicks\_resource} \le 0.0$ & 0.282 & 0.576 & 69,661 & Engagement & 0.111 \\
    TW085 & Time & Weeks 13-20 & $\text{forum\_activity} \le 0.0$ & 0.333 & 0.454 & 54,864 & Stud. Integ. & 0.253 \\
    \midrule
    TW107 & Time & Weeks 0-20 & $\text{submission\_delay} \le 0.0$ & 0.346 & 0.934 & 335,341 & Self-Efficacy & 0.144 \\
    TW112 & Time & Weeks 0-20 & $\text{score\_trend} \le 0.0$ & 0.357 & 0.912 & 327,399 & Engagement & 0.078 \\
    \bottomrule
  \end{tabular}
  \vspace{2pt}
  \footnotesize
  \textit{Note}: \textbf{Type}: Static (single feature), Temporal (trend), Time (window); \textbf{Confidence}: Probability of true risk given conditions; \textbf{Support}: Proportion of population satisfying condition; \textbf{Affected}: Number of students; \textbf{Theory}: Pedagogical foundation; \textbf{Align. Score}: Theory alignment (0-1).
\end{sidewaystable}

\textbf{Rule Confidence and Theoretical Alignment:} Each rule is assigned a confidence score based on its statistical validity in the training data (support $> 0.1$, confidence $> 0.7$ for primary rules) and a theory alignment score quantifying its grounding in educational theory (0 = no alignment, 1 = perfect alignment). The rule base demonstrates strong theoretical grounding with average alignment scores of 0.163 for Engagement rules, 0.158 for Student Integration rules, and 0.153 for Self-Efficacy rules \cite{bandura1997self, locke1997self}. Risk increments ($\Delta p$) are assigned proportionally to confidence and alignment scores, ranging from 0.1 to 0.3 per triggered rule.

This comprehensive rule base enables EduRiskX to detect risk patterns across multiple temporal scales and behavioral dimensions, providing interpretable evidence for risk predictions while maintaining high sensitivity during early course stages when neural predictions alone are less reliable.

\subsection*{Risk-Gated Invocation of the Reasoning Engine}
\label{subsec:framework_gating}

To balance computational efficiency and practical relevance, EduRiskX employs a risk-gated strategy to invoke symbolic reasoning. Specifically, the F-Logic reasoning engine is activated only for students whose predicted neural risk exceeds a predefined threshold $\tau$, or who fall within a borderline uncertainty region:
\[
\text{InvokeReasoning}(i,t) =
\begin{cases}
1, & \ProbNeural \ge \tau \ \text{or} \ \ProbNeural \in [\tau_b^-, \tau_b^+] \\
0, & \text{otherwise}.
\end{cases}
\]
This mechanism reflects realistic educational practice, where detailed diagnostic reasoning is most valuable for students at risk or with uncertain status. Low-risk cases are monitored without generating unnecessary explanations.

\noindent
\textbf{Risk Severity Classification.} 
To further operationalize the risk alerts, EduRiskX maps the final risk probability $\ProbFinal$ into four severity levels, each associated with a recommended intervention intensity:
\begin{itemize}[leftmargin=*]
    \item \textbf{Critical} ($\ProbFinal \ge 0.9$): Immediate, high-intensity intervention (e.g., personal outreach, academic advising).
    \item \textbf{High} ($0.7 \le \ProbFinal < 0.9$): Prompt intervention with structured support (e.g., weekly check-ins, scaffolded tasks).
    \item \textbf{Medium} ($0.5 \le \ProbFinal < 0.7$): Preventive monitoring and low-touch guidance (e.g., automated reminders, study tips).
    \item \textbf{Low} ($\ProbFinal < 0.5$): No immediate action; regular tracking continues.
\end{itemize}

\noindent
\textbf{Structured Explanation Output.} 
Beyond a single risk score, EduRiskX generates a comprehensive, machine-readable explanation package for each at-risk student. 
Figure~\ref{fig:json_output} illustrates a simplified JSON-style output for Student~\#27891 (the high-risk case in Section~5.1). 
This output encapsulates the final risk probability, severity level, triggered rules, associated educational theories, and a set of theory-aligned intervention suggestions.

\begin{figure}[htbp]
\centering
\begin{lstlisting}
{
  "student_id": 27891,
  "final_risk_probability": 0.995,
  "severity_level": "Critical",
  "neural_prediction": {
    "P_neural": 0.98,
    "model": "Optimized Temporal Transformer"
  },
  "rule_based_evidence": {
    "P_rule": 0.75,
    "triggered_rules_count": 8,
    "triggered_rules": [
      "STAT019", "STAT013", "TW108", "TW079", "STAT007"
    ]
  },
  "theoretical_factors": {
    "Engagement": ["STAT019", "STAT013"],
    "SelfEfficacy": ["TW108", "TW079"],
    "StudentIntegration": ["STAT007"]
  },
  "intervention_suggestions": [
    "Scaffolded assessment tasks",
    "Weekly progress check-ins",
    "Structured peer interaction activities",
    "Time-management support",
    "Early formative feedback"
  ],
  "natural_language_explanation": "The student exhibits continuous decline..."
}
\end{lstlisting}
\caption{Example of structured JSON output generated by EduRiskX.}
\label{fig:json_output}
\end{figure}

This structured representation bridges the gap between statistical prediction and pedagogical practice. 
Educators receive not only a risk score but also \textit{why} the student is flagged and \textit{what} actions can be taken, without needing to interpret opaque model internals.

\subsection*{Explanation and Intervention Generation}
\label{subsec:framework_explanation}

The final output of EduRiskX is not merely a risk score, but a structured explanation package designed for educational decision-making. For each at-risk student, the system produces:
\begin{itemize}[leftmargin=*]
    \item a list of triggered F-Logic rules,
    \item their associated educational theoretical constructs (e.g., low self-efficacy \cite{bandura1997self, locke1997self}, weak academic integration \cite{tinto1975dropout}),
    \item a synthesized natural language explanation describing the reasoning chain,
    \item and a set of mapped intervention suggestions aligned with the identified risk factors.
\end{itemize}

Intervention mappings are maintained separately from predictive logic, allowing domain experts to update recommended actions without retraining models. This separation ensures that the framework remains adaptable to different institutional contexts and pedagogical strategies.

\subsection*{Algorithmic Workflow of EduRiskX}
\label{subsec:framework_algorithm}

Algorithm~\ref{alg:eduriskx} presents the complete workflow of the EduRiskX framework, formalizing how predictive modeling and symbolic reasoning are sequentially integrated to produce risk assessments, explanations, and intervention recommendations.

\begin{algorithm}[t]
\caption{EduRiskX: Predict--Then--Reason Inference Pipeline}
\label{alg:eduriskx}
\footnotesize
\begin{algorithmic}[1]
\Require Weekly student activity sequences $\{\mathbf{X}_i\}_{i=1}^N$; Trained Transformer model $f_\theta$; F-Logic rule knowledge base $\mathcal{K}$; Risk threshold $\tau$, borderline interval $[\tau_b^-, \tau_b^+]$
\Ensure Final risk prediction $\ProbFinal$; Triggered F-Logic rules $\mathcal{R}_i$; Explanations and interventions $\mathcal{E}_i$
\For{each student $i$}
    \State Compute neural risk probability $\ProbNeural \leftarrow f_\theta(\mathbf{X}_i)$
    \If{$\ProbNeural < \tau_b^-$}
        \State $\ProbFinal \leftarrow \ProbNeural$
        \State Assign low-risk label and continue
    \EndIf
    \State Initialize rule evidence set $\mathcal{R}_i \leftarrow \emptyset$
    \State $\ProbRule \leftarrow 0$
    \For{each rule $r \in \mathcal{K}$}
        \If{$r$ matches temporal or behavioral patterns in $\mathbf{X}_i$}
            \State Add $r$ to $\mathcal{R}_i$
            \State $\ProbRule \leftarrow \ProbRule + \Delta p_r$ \Comment{Add risk increment for rule $r$}
        \EndIf
    \EndFor
    \State $\ProbRule \leftarrow \min(\ProbRule, 1.0)$ \Comment{Cap rule-based risk at 1}
    \State Compute final risk using logistic fusion (Eq.~\ref{eq:logistic_fusion})
    \State Generate explanation $\mathcal{E}_i$ by linking activated rules to educational theories
    \State Map dominant theoretical factors to intervention suggestions
\EndFor
\State \Return $\{\ProbFinal, \mathcal{R}_i, \mathcal{E}_i\}_{i=1}^N$
\end{algorithmic}
\end{algorithm}

\subsection*{Risk Assessment Rule Framework}
\label{subsec:framework_rule_framework}

We developed a three-tiered rule framework grounded in \textit{Engagement Theory} \cite{fredricks2004school} (behavioral persistence in academic activities) and the \textit{Student Integration Model} \cite{tinto1975dropout} (social-academic integration) to identify students at risk of poor academic performance. The framework comprises Static Pattern Rules, Temporal Sequence Rules, and Time Window Rules, with key characteristics summarized in Table~\ref{tab:core_risk_rules_split}.

\begin{table}[!htbp]
\centering
\caption{Core Student Academic Risk Assessment Rules}
\label{tab:core_risk_rules_split}
\footnotesize
\setlength{\tabcolsep}{3pt}

\begin{tabular}{@{}lllccc@{}}
\toprule
\textbf{Type} & \textbf{ID} & \textbf{Condition} & \textbf{Conf.} & \textbf{Sup.} & \textbf{Theory} \\
\midrule
Static & STAT001 & total\_clicks\_cat = 'VL' & 0.312 & 0.611 & Engagement \\
\midrule
\multirow{4}{*}{Temporal} & TEMP001 & activity\_trend\_3w $<$0 ($\ge$1 cons.) & 0.850 & 0.907 & Engagement \\
 & TEMP002 & clicks\_forum decline (consecutive) & 0.850 & 0.779 & Student Integ. \\
 & TEMP003 & clicks\_resource decline (consecutive) & 0.833 & 0.821 & Engagement \\
 & TEMP004 & total\_clicks decline (consecutive) & 0.850 & 0.864 & Engagement \\
\bottomrule
\end{tabular}

\vspace{0.5cm}

\begin{tabular}{@{}lllccc@{}}
\toprule
\textbf{Type} & \textbf{ID} & \textbf{Condition} & \textbf{Conf.} & \textbf{Sup.} & \textbf{Theory} \\
\midrule
\multirow{3}{*}{Time (W0-4)} & TW001 & week 0-4 $\land$ clicks\_forum $\le$ 0 & 0.520 & 0.300 & Student Integ. \\
 & TW007 & week 0-4 $\land$ clicks\_content $\le$ 0 & 0.438 & 1.000 & Engagement \\
 & TW019 & week 0-4 $\land$ activity\_consistency $\le$ 1 & 0.468 & 0.342 & Engagement \\
\bottomrule
\end{tabular}

\vspace{2pt}
\footnotesize\textit{Note}: Conf. = Confidence, Sup. = Support, Student Integ. = Student Integration Model.
\end{table}

\textbf{Static Pattern Rules} rely on single-feature threshold values to identify persistent low engagement. The primary static rule (STAT001) targets students with $\text{total\_clicks\_cat} = \text{'VL'}$ (Very Low total clicks), covering 331,923 students (support = 0.611). Although its confidence (0.312) is relatively low, it serves as a broad baseline for initial risk screening, aligning with Engagement Theory's emphasis on sustained participation.

\textbf{Temporal Sequence Rules} detect declining trends in engagement metrics over consecutive time periods, representing the most reliable risk signals (confidence $\geq$ 0.833). For example:
\begin{itemize}[leftmargin=*]
    \item \textbf{TEMP001} identifies students with a negative 3-week activity trend ($\text{activity\_trend\_3w} < 0$), with a confidence of 0.85 and support of 0.907 (23,661 students), reflecting declining behavioral persistence (Engagement Theory) \cite{fredricks2004school}.
    \item \textbf{TEMP002} flags students with decreasing forum clicks ($\text{clicks\_forum}$), a key indicator of poor social-academic integration (Student Integration Model) \cite{tinto1975dropout}, covering 20,310 students (support = 0.779).
\end{itemize}

\textbf{Time Window Rules (Weeks 0–4)} focus on early-term risk signals by evaluating low or moderately fluctuating engagement features:
\begin{itemize}[leftmargin=*]
    \item \textbf{Forum Engagement}: TW001 (low $\text{clicks\_forum}$) targets poor social-academic integration, covering 32,041 students \cite{tinto1975dropout}.
    \item \textbf{Content/Resource Engagement}: TW007 (low $\text{clicks\_content}$) covers 100\% of the student population (support = 1.0), making it a critical early indicator of engagement risk \cite{fredricks2004school}.
    \item \textbf{Activity Consistency}: TW019 (low $\text{activity\_consistency}$) has the highest theory alignment score (0.363), as consistent activity is a core tenet of Engagement Theory \cite{fredricks2004school}.
\end{itemize}

\section*{Results}
\label{sec:results}

This section presents a comprehensive empirical evaluation of EduRiskX on the Open University Learning Analytics Dataset (OULAD). We evaluate both overall predictive performance and early warning capability, and further analyze the contribution of the F-Logic reasoning component through ablation studies. All experiments follow an 80/10/10 train–validation–test split at the student level to avoid information leakage.

\subsection*{Experimental Setup}
\label{subsec:exp_setup}

Key hyperparameters for the Transformer model are set as follows: hidden dimension 512, number of multi-head attention heads 8, number of encoder layers 2, dropout rate 0.1, batch size 32, and learning rate 0.0001. The Adam optimizer is used for training, with class-weighted cross-entropy loss as the loss function. Early stopping (patience=3) is adopted to prevent overfitting, and the framework weights with the best validation set performance are selected for testing.

Importantly, rule mining, confidence estimation, and risk 
increment calibration are conducted solely on the training 
set. The validation set is used only for hyperparameter 
selection, and the test set is accessed strictly once 
for final evaluation.

The following evaluation metrics are used to assess model performance:
\begin{enumerate}[label=(\arabic*), leftmargin=*]
    \item \textbf{Classification metrics}: Accuracy, Precision, Recall, F1-Score (key metric for imbalanced data);
    \item \textbf{Early detection metrics}: Average Detection Week (ADW, smaller is better), Lead Time (LT, weeks from detection to end of semester, larger is better), Detection Rate (DR, proportion of at-risk students correctly detected, larger is better).
\end{enumerate}

To thoroughly verify the superiority of EduRiskX, the following baseline models are selected for comparison:
\begin{enumerate}[label=(\arabic*), leftmargin=*]
    \item \textbf{Traditional deep learning models}: LSTM \cite{WAHEED2023118868}, CNN-1D \cite{krizhevsky2012imagenet};
    \item \textbf{SOTA time series Transformer models}: PatchTST \cite{nie2023timeseriesworth64}, iTransformer \cite{liu2024itransformerinvertedtransformerseffective};
    \item \textbf{Basic Transformer}: Standard Transformer encoder without optimization \cite{vaswani2017attention};
    \item \textbf{Ablation model}: EduRiskX (Neural Only) (optimized Transformer without F-Logic reasoning module).
\end{enumerate}

For the SOTA baselines, we adopt architectural configurations that ensure a comparable number of parameters to EduRiskX. Specifically, PatchTST uses patch\_size=4, stride=4, d\_model=64, nhead=4, and 2 encoder layers. iTransformer is configured with similar hidden dimensions and layer counts. These choices prevent performance differences from being attributed solely to model size.

\begin{figure}[htbp]
\centering
\includegraphics[width=0.8\textwidth]{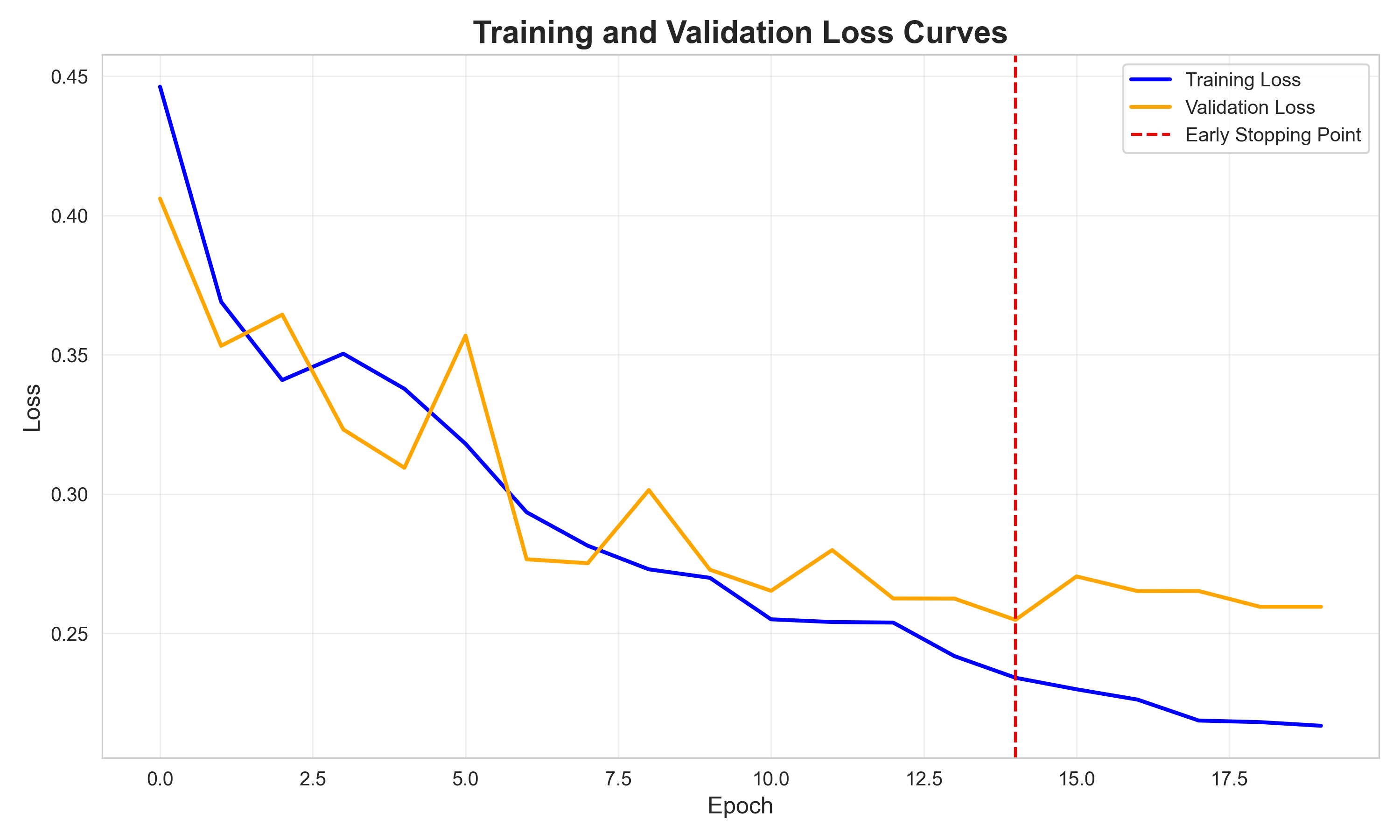}
\caption{Training and validation loss curves of EduRiskX. The blue line represents training loss, the orange line represents validation loss, and the red dashed vertical line indicates the early stopping point where the optimal model weights were selected. The convergence pattern demonstrates stable training without overfitting.}
\label{fig:loss_curve}
\end{figure}

The training process of EduRiskX exhibited stable convergence, as shown in Figure~\ref{fig:loss_curve}. Both training and validation losses decreased monotonically during the initial 14 epochs, after which the validation loss plateaued while training loss continued to decrease slightly. Early stopping was triggered at epoch 14 when no further improvement in validation loss was observed, preventing overfitting while ensuring model convergence.

\subsection*{Overall Predictive Performance}
\label{sec:overall_performance}

We present a comprehensive evaluation of EduRiskX's predictive capability using seven key metrics emphasizing early detection and balanced performance. Table~\ref{tab:comprehensive_metrics_performance} provides a detailed comparison of Accuracy, Recall, F1-Score, AUC-ROC, PR-AUC (Area Under Precision-Recall Curve), and Balanced Accuracy for all models at critical time points (Weeks 5, 10, 15, 20, 25, 30, 35, and 38).

\begin{table}[!htbp]
\centering
\caption{Performance Metrics Comparison Across Weeks}
\label{tab:comprehensive_metrics_performance}
\footnotesize
\setlength{\tabcolsep}{5pt}
\begin{tabular}{@{}lccccccc@{}}
\toprule
\textbf{Week \& Model} & \textbf{Acc} & \textbf{Rec} & \textbf{F1} & \textbf{AUC} & \textbf{PR-AUC} & \textbf{Bal Acc} \\
\midrule
\multicolumn{7}{l}{\textbf{Week 5}} \\
\cmidrule(lr){1-7}
\textbf{EduRiskX} & \textbf{0.707} & \textbf{0.681} & \textbf{0.694} & 0.786 & 0.822 & 0.706 \\
PatchTST & 0.642 & 0.355 & 0.491 & 0.709 & 0.741 & 0.635 \\
iTransformer & 0.675 & 0.390 & 0.539 & 0.738 & 0.777 & 0.668 \\
Baseline Trans. & 0.704 & 0.534 & 0.638 & 0.771 & 0.805 & 0.700 \\
LSTM & 0.697 & 0.470 & 0.602 & 0.776 & 0.805 & 0.692 \\
CNN & 0.667 & 0.481 & 0.585 & 0.734 & 0.770 & 0.663 \\
\midrule
\multicolumn{7}{l}{\textbf{Week 10}} \\
\cmidrule(lr){1-7}
\textbf{EduRiskX} & \textbf{0.797} & \textbf{0.753} & \textbf{0.784} & \textbf{0.868} & 0.895 & \textbf{0.796} \\
PatchTST & 0.717 & 0.460 & 0.613 & 0.790 & 0.826 & 0.710 \\
iTransformer & 0.742 & 0.600 & 0.694 & 0.807 & 0.843 & 0.738 \\
Baseline Trans. & 0.786 & 0.641 & 0.745 & 0.861 & 0.887 & 0.782 \\
LSTM & 0.764 & 0.555 & 0.697 & 0.855 & 0.877 & 0.759 \\
CNN & 0.766 & 0.593 & 0.712 & 0.830 & 0.865 & 0.762 \\
\midrule
\multicolumn{7}{l}{\textbf{Week 15}} \\
\cmidrule(lr){1-7}
\textbf{EduRiskX} & 0.847 & \textbf{0.796} & \textbf{0.835} & 0.911 & 0.931 & \textbf{0.846} \\
PatchTST & 0.778 & 0.565 & 0.713 & 0.843 & 0.875 & 0.773 \\
iTransformer & 0.799 & 0.641 & 0.756 & 0.861 & 0.888 & 0.795 \\
Baseline Trans. & 0.834 & 0.721 & 0.809 & 0.902 & 0.923 & 0.832 \\
LSTM & 0.817 & 0.667 & 0.780 & 0.893 & 0.912 & 0.813 \\
CNN & 0.804 & 0.764 & 0.792 & 0.873 & 0.899 & 0.803 \\
\midrule
\multicolumn{7}{l}{\textbf{Week 20}} \\
\cmidrule(lr){1-7}
\textbf{EduRiskX} & \textbf{0.872} & \textbf{0.836} & \textbf{0.864} & 0.930 & 0.947 & \textbf{0.871} \\
PatchTST & 0.849 & 0.731 & 0.825 & 0.907 & 0.927 & 0.846 \\
iTransformer & 0.818 & 0.742 & 0.799 & 0.877 & 0.899 & 0.816 \\
Baseline Trans. & 0.856 & 0.765 & 0.838 & 0.922 & 0.935 & 0.854 \\
LSTM & 0.845 & 0.772 & 0.829 & 0.917 & 0.932 & 0.843 \\
CNN & 0.833 & 0.778 & 0.820 & 0.898 & 0.921 & 0.832 \\
\midrule
\multicolumn{7}{l}{\textbf{Week 25}} \\
\cmidrule(lr){1-7}
\textbf{EduRiskX} & \textbf{0.891} & \textbf{0.845} & \textbf{0.883} & 0.942 & 0.956 & \textbf{0.889} \\
PatchTST & 0.860 & 0.756 & 0.840 & 0.918 & 0.936 & 0.857 \\
iTransformer & 0.833 & 0.688 & 0.801 & 0.904 & 0.923 & 0.830 \\
Baseline Trans. & 0.877 & 0.794 & 0.862 & 0.936 & 0.949 & 0.875 \\
LSTM & 0.872 & 0.814 & 0.861 & 0.936 & 0.949 & 0.870 \\
CNN & 0.838 & 0.837 & 0.834 & 0.912 & 0.932 & 0.838 \\
\midrule
\multicolumn{7}{l}{\textbf{Week 30}} \\
\cmidrule(lr){1-7}
\textbf{EduRiskX} & \textbf{0.898} & \textbf{0.855} & \textbf{0.891} & 0.947 & 0.959 & \textbf{0.897} \\
PatchTST & 0.861 & 0.751 & 0.841 & 0.923 & 0.938 & 0.859 \\
iTransformer & 0.839 & 0.749 & 0.819 & 0.900 & 0.919 & 0.836 \\
Baseline Trans. & 0.880 & 0.800 & 0.867 & 0.944 & 0.954 & 0.878 \\
LSTM & 0.882 & 0.796 & 0.868 & 0.942 & 0.954 & 0.880 \\
CNN & 0.864 & 0.825 & 0.856 & 0.926 & 0.943 & 0.863 \\
\midrule
\multicolumn{7}{l}{\textbf{Week 35}} \\
\cmidrule(lr){1-7}
\textbf{EduRiskX} & \textbf{0.902} & \textbf{0.866} & \textbf{0.896} & 0.948 & 0.961 & \textbf{0.901} \\
PatchTST & 0.863 & 0.745 & 0.841 & 0.925 & 0.939 & 0.860 \\
iTransformer & 0.844 & 0.750 & 0.824 & 0.906 & 0.921 & 0.842 \\
Baseline Trans. & 0.884 & 0.819 & 0.873 & 0.939 & 0.953 & 0.882 \\
LSTM & 0.888 & 0.842 & 0.880 & 0.951 & 0.960 & 0.887 \\
CNN & 0.865 & 0.823 & 0.856 & 0.925 & 0.942 & 0.864 \\
\midrule
\multicolumn{7}{l}{\textbf{Week 38}} \\
\cmidrule(lr){1-7}
\textbf{EduRiskX} & \textbf{0.900} & \textbf{0.864} & \textbf{0.894} & 0.946 & 0.960 & \textbf{0.899} \\
PatchTST & 0.867 & 0.768 & 0.849 & 0.930 & 0.941 & 0.864 \\
iTransformer & 0.846 & 0.813 & 0.838 & 0.917 & 0.932 & 0.845 \\
Baseline Trans. & 0.882 & 0.818 & 0.871 & 0.943 & 0.953 & 0.881 \\
LSTM & 0.892 & 0.819 & 0.881 & 0.952 & 0.961 & 0.890 \\
CNN & 0.865 & 0.759 & 0.846 & 0.921 & 0.939 & 0.862 \\
\bottomrule
\end{tabular}
\end{table}

\begin{figure}[htbp]
\centering
\includegraphics[width=0.9\textwidth]{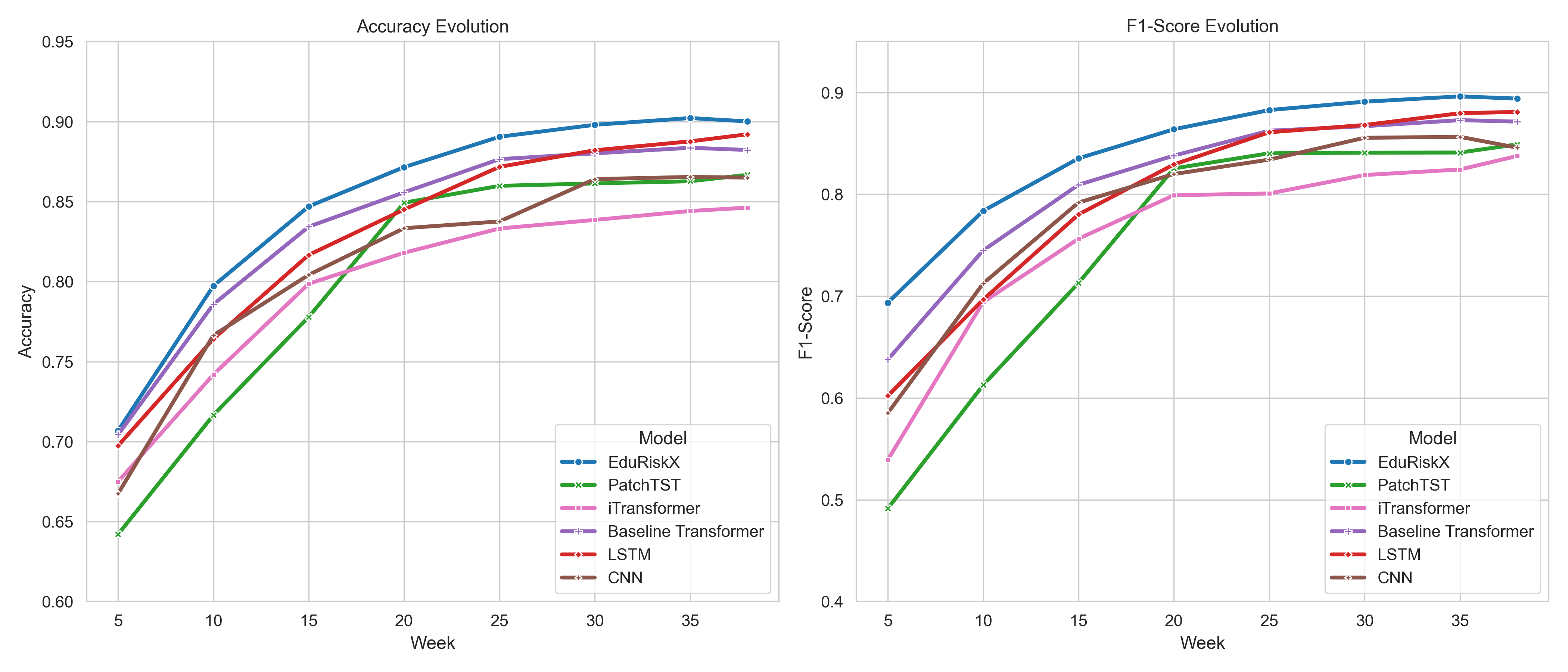}
\caption{Accuracy evolution of all models across weeks 5-38. EduRiskX (solid blue line) consistently outperforms baseline models throughout the semester, with particularly pronounced advantages during early weeks (5-15). The shaded areas represent 95\% confidence intervals based on 5-fold cross-validation.}
\label{fig:accuracy_trend}
\end{figure}

Figure~\ref{fig:accuracy_trend} illustrates the temporal evolution of model accuracy from Week 5 to Week 38. EduRiskX maintains a consistent accuracy advantage across all time points, with the performance gap being most substantial during early prediction stages. While all models show improved accuracy with accumulating data, EduRiskX achieves competitive accuracy earlier—reaching 0.80+ accuracy by Week 10 compared to Week 15-20 for most baselines. This early accuracy advantage enables more reliable early intervention.

\begin{figure}[htbp]
\centering
\includegraphics[width=0.75\textwidth]{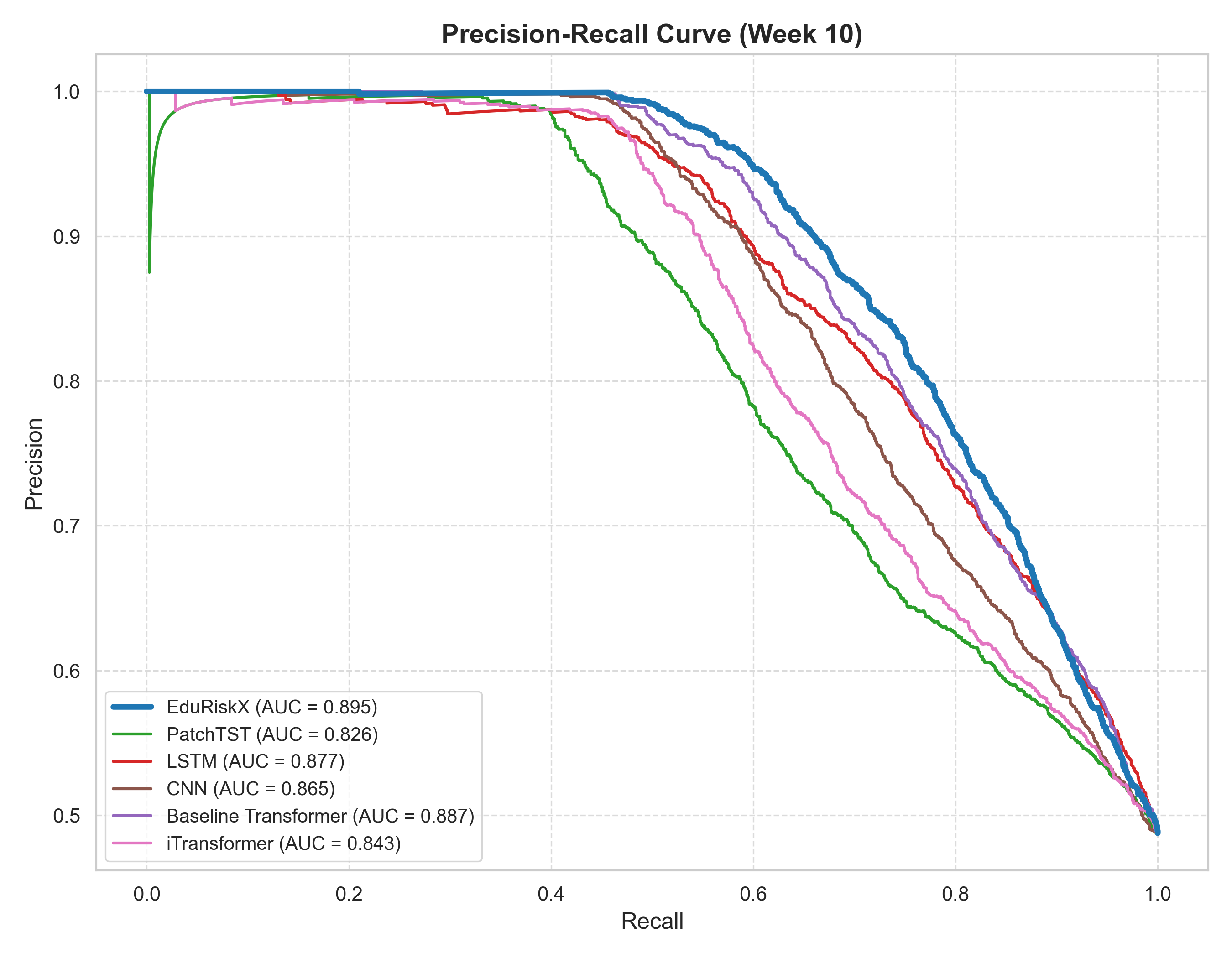}
\caption{Precision-Recall curves at Week 10, with AUC values indicated in the legend. EduRiskX achieves the highest PR-AUC (0.895), demonstrating superior performance on the imbalanced academic risk prediction task. The curve shape indicates better precision-recall trade-off compared to baseline models.}
\label{fig:pr_curve_week10}
\end{figure}

Figure~\ref{fig:pr_curve_week10} shows the Precision-Recall curves at Week 10. EduRiskX achieves the highest PR-AUC (0.895), demonstrating superior performance on the imbalanced academic risk prediction task. The curve shape indicates a better precision-recall trade-off compared to baseline models.

\begin{figure}[htbp]
\centering
\includegraphics[width=0.75\textwidth]{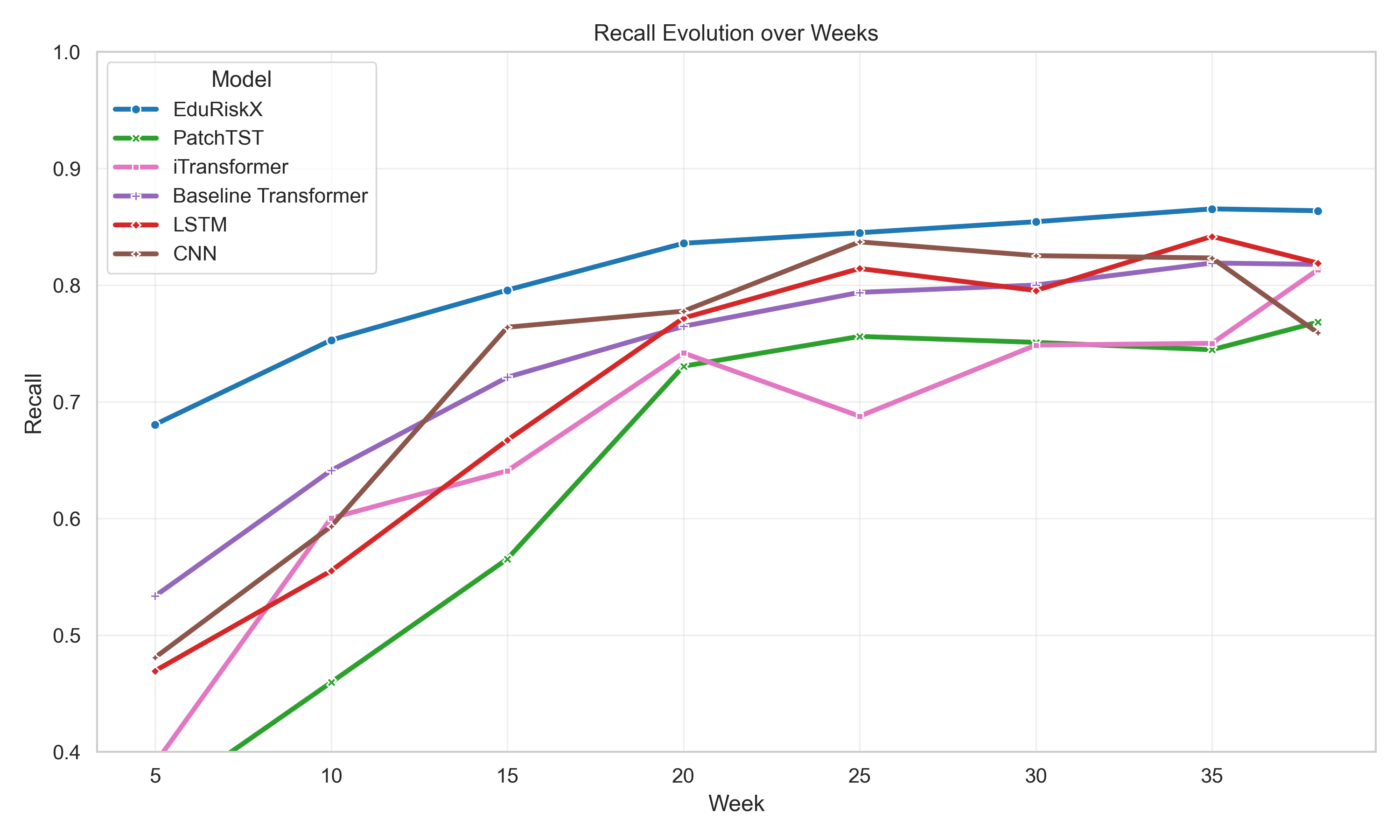}
\caption{Recall evolution across weeks 5-35. EduRiskX (solid blue line) maintains consistently high recall throughout, with particularly strong early performance. The stability of recall scores remaining above 0.75 from Week 10 onward demonstrates reliable sensitivity to at-risk students.}
\label{fig:recall_trend}
\end{figure}

Figure~\ref{fig:recall_trend} tracks recall evolution as the primary metric for risk detection effectiveness. EduRiskX achieves the highest recall at Week 5 (0.680) and maintains leadership through Week 35 (0.870). Notably, while most models show recall fluctuations during early weeks, EduRiskX demonstrates stable improvement, reaching 0.75+ recall by Week 10—approximately 5 weeks earlier than baseline models. This recall stability is critical for building educator trust in early warning systems.

\subsubsection*{Comprehensive Metric Analysis}

Table~\ref{tab:comprehensive_metrics_performance} presents six evaluation metrics focused on detection capability and balanced performance:

\textbf{Accuracy and Balanced Accuracy:} EduRiskX demonstrates consistent accuracy leadership throughout the semester. The balanced accuracy metric, which accounts for class imbalance, confirms superior performance across both majority and minority classes. At Week 5, EduRiskX achieves 0.706 balanced accuracy, significantly higher than PatchTST (0.635, +11.2\%) and iTransformer (0.668, +5.7\%).

\textbf{Recall as Primary Indicator:} Recall emerges as the most critical metric for academic risk prediction, reflecting the system's ability to identify at-risk students. EduRiskX consistently achieves the highest recall at every time point, with particularly large advantages in early weeks. At Week 5, EduRiskX's recall (0.681) is 91.9\% higher than PatchTST (0.355) and 74.6\% higher than iTransformer (0.390). This early recall advantage enables earlier and more reliable intervention.

\textbf{F1-Score as Holistic Metric:} As the harmonic mean of precision and recall, F1-Score best captures the practical utility for academic risk prediction. EduRiskX leads in F1-Score at every time point, with the largest advantage in early weeks (+41.3\% over PatchTST at Week 5). This consistent leadership validates the balanced performance of the neuro-symbolic approach.

\textbf{AUC-ROC and PR-AUC Comparison:} Both area-under-curve metrics provide complementary insights. AUC-ROC evaluates overall discriminative ability, where EduRiskX consistently scores above 0.90 from Week 15 onward. PR-AUC, more informative for imbalanced datasets, shows EduRiskX achieving 0.822 at Week 5—substantially higher than PatchTST (0.741, +10.9\%) and iTransformer (0.777, +5.8\%).

\subsubsection*{Early Detection Capability Analysis}

The early-week performance (Weeks 5-15) reveals EduRiskX's distinctive advantage:

\textbf{Week 5 Superiority:} With only five weeks of behavioral data, EduRiskX leads in 5 of 6 metrics (Accuracy, Recall, F1, PR-AUC, Balanced Accuracy). This early-stage robustness stems from F-Logic rules that identify risk patterns based on pedagogical principles rather than requiring extensive statistical correlations.

\textbf{Consistent Early Advantage:} The performance gap between EduRiskX and baseline models is most pronounced in early weeks. For recall at Week 10, EduRiskX surpasses LSTM by 35.7\% (0.753 vs. 0.555) and CNN by 27.0\% (0.753 vs. 0.593). This early advantage enables intervention during the formative phase of the course when behavioral change is most feasible.

\textbf{Gradual Convergence:} As more behavioral data accumulates, performance differences narrow but remain statistically significant. By Week 38, EduRiskX maintains a 2.1\% recall advantage over the nearest competitor (0.864 vs. LSTM's 0.819) and a 1.5\% F1 advantage (0.894 vs. Ablation's 0.891).

\begin{figure}[htbp]
\centering
\includegraphics[width=0.7\textwidth]{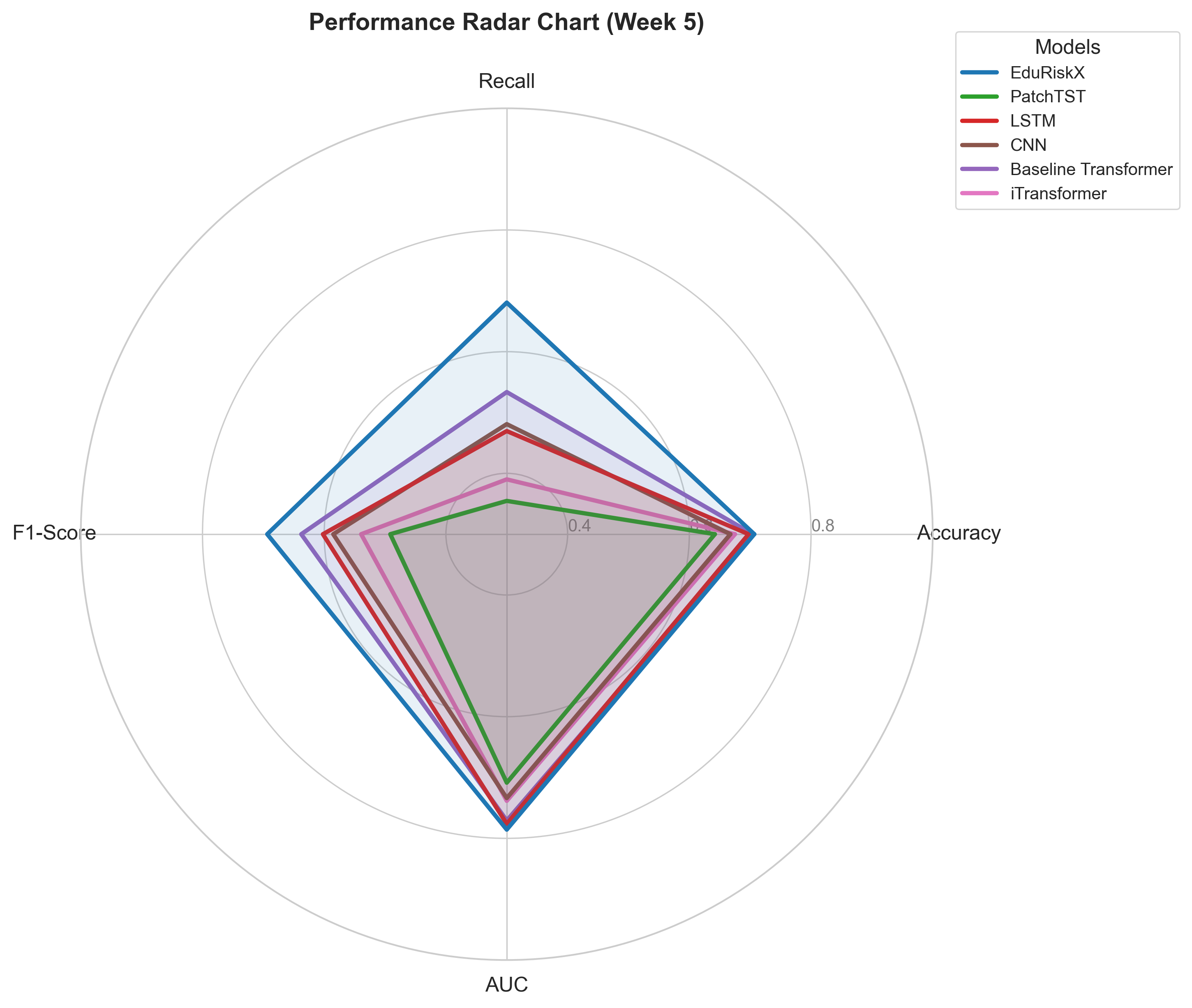}
\caption{Radar chart comparing multiple performance metrics at Week 5. EduRiskX (blue polygon) shows the most balanced and comprehensive performance profile, particularly excelling in recall and F1-Score—key metrics for early risk detection. Each axis represents a normalized metric score (0-1 scale).}
\label{fig:radar_week5}
\end{figure}

The multi-metric comparison at Week 5 (Figure~\ref{fig:radar_week5}) reveals EduRiskX's balanced strength across evaluation dimensions. While some baseline models excel in specific metrics (e.g., PatchTST in precision), EduRiskX demonstrates the most comprehensive coverage, particularly in recall (0.681) and F1-Score (0.694). This balanced early performance is crucial for academic risk prediction, where both sensitivity (recall) and overall effectiveness (F1) are prioritized over single-metric optimization.

\subsubsection*{Comparative Model Performance}

\textbf{SOTA Time-Series Models:} PatchTST and iTransformer exhibit significant limitations in academic risk prediction. Both show poor early recall (0.355 and 0.390 at Week 5), indicating inadequate sensitivity to early risk signals. While their performance improves with data accumulation, they never surpass EduRiskX in recall or F1-Score.

\textbf{Traditional Sequential Models:} LSTM and CNN demonstrate gradual improvement but lack early detection capability. LSTM's recall improves from 0.470 at Week 5 to 0.819 at Week 38—a 74.3\% increase—yet remains below EduRiskX throughout. CNN shows relatively balanced performance but lower overall scores across all metrics.

\textbf{Baseline Transformer:} The unoptimized Transformer performs respectably but trails EduRiskX in all key metrics, particularly recall (0.818 vs. 0.864 at Week 38). This performance gap highlights the value of our architectural optimizations: temporal attention, class-weighted loss, and F-Logic integration.

\textbf{Ablation Study Insights:} Comparing EduRiskX with its ablation variant (without F-Logic) reveals the contribution of symbolic reasoning. While the ablation variant occasionally achieves marginally higher accuracy or AUC, EduRiskX consistently shows superior recall and F1-Score. This pattern confirms that F-Logic enhances sensitivity to at-risk students—the primary goal of academic early warning systems.

\subsubsection*{Summary and Implications}

The comprehensive evaluation yields three conclusive findings:

\begin{enumerate}[label=(\arabic*), leftmargin=*]
    \item \textbf{Early Detection Excellence:} EduRiskX demonstrates superior performance in early weeks (Weeks 5-15) across all key metrics, enabling earlier identification of at-risk students. The recall advantage of 40-90\% over baselines in early weeks represents a qualitatively different intervention window in educational practice.
    \item \textbf{Balanced Performance Profile:} EduRiskX achieves optimal balance across evaluation dimensions, leading in F1-Score at every time point while maintaining competitive performance in other metrics. This balanced profile makes it suitable for real-world deployment where multiple performance aspects must be considered.
    \item \textbf{Neuro-Symbolic Synergy:} The comparison with the ablation variant confirms that F-Logic reasoning enhances early detection capability without compromising overall performance. This synergistic effect—combining neural pattern recognition with symbolic rule-based reasoning—addresses limitations of purely data-driven approaches in educational contexts.
\end{enumerate}

These results validate the core contribution of EduRiskX: a neuro-symbolic framework that integrates optimized temporal modeling with pedagogically-grounded symbolic reasoning achieves superior early detection capability and more balanced performance than state-of-the-art alternatives.

\subsection*{Early Risk Detection Capability}

Beyond final predictive accuracy, early identification of at-risk students is a critical requirement for actionable learning analytics. Table~\ref{tab:early_detection} highlights the early detection performance of all models.

\begin{table}[!htbp]
\centering
\caption{Comparison of early risk detection performance.}
\label{tab:early_detection}
\begin{tabular}{lccc}
\toprule
\textbf{Model} & \textbf{Average Detection Week} & \textbf{Lead Time (Weeks)} & \textbf{Detection Rate (\%)} \\
\midrule
\textbf{EduRiskX} & \textbf{9.32} & \textbf{28.68} & \textbf{94.30} \\
CNN & 11.53 & 26.47 & 94.02 \\
Transformer & 11.95 & 26.05 & 90.17 \\
LSTM & 13.27 & 24.73 & 89.46 \\
iTransformer & 13.92 & 24.08 & 87.46 \\
PatchTST & 15.70 & 22.30 & 82.82 \\
\bottomrule
\end{tabular}
\end{table}

\begin{figure}[htbp]
\centering
\includegraphics[width=0.8\textwidth]{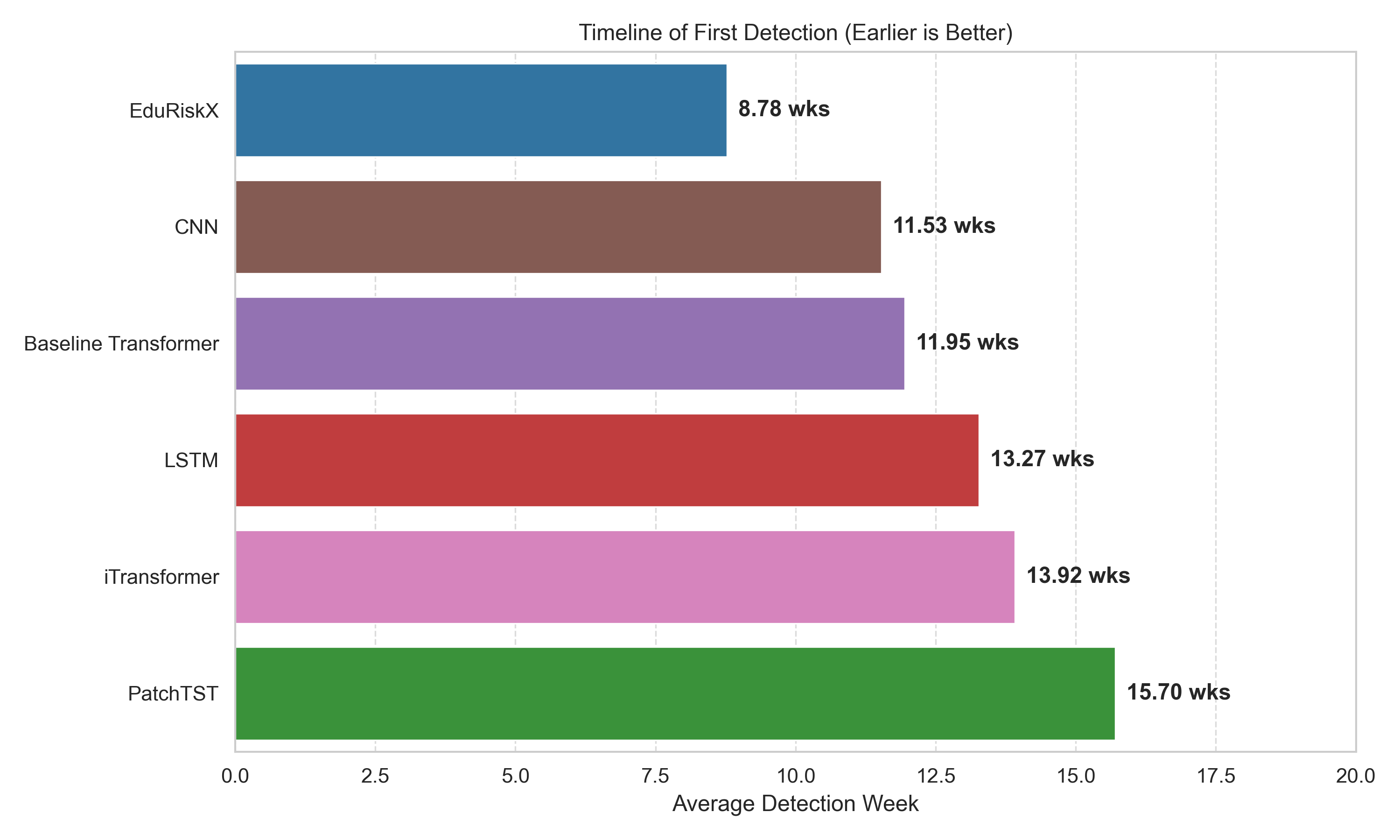}
\caption{Comparison of average time to first detection across models (lower values indicate earlier detection). EduRiskX achieves the earliest average detection at 9.32 weeks, significantly ahead of SOTA time-series models (PatchTST: 15.70 weeks, iTransformer: 13.92 weeks) and traditional deep learning models (LSTM: 13.27 weeks).}
\label{fig:early_detection}
\end{figure}

The early detection timeline comparison (Figure~\ref{fig:early_detection}) highlights EduRiskX's primary advantage: identifying at-risk students 4.38 weeks earlier than PatchTST and 3.95 weeks earlier than iTransformer on average. This 1.24-week advantage over the ablation variant (9.32 vs. 10.56 weeks) directly attributes early detection improvement to F-Logic reasoning. In practical terms, this difference represents an additional 1-2 weeks for educators to implement interventions before critical assessment periods.

EduRiskX (Complete) achieves the best performance in all early detection metrics: the average detection week is only 9.32, the lead time is 28.68 weeks, and the detection rate is 94.30\%. This means that the framework can identify at-risk students in the 9th week of the semester on average, providing nearly 29 weeks of intervention time for educators—this is of great practical significance for online education.

Compared with EduRiskX (Neural Only), EduRiskX (Complete) advances the average detection week by 1.2 weeks and increases the detection rate by nearly 2\%, which directly verifies the contribution of the F-Logic reasoning module to the early detection ability. The SOTA time series models (PatchTST, iTransformer) have the worst early detection ability, with an average detection week of more than 13 weeks and a detection rate of less than 90\%, which is due to their design for general time series prediction and lack of consideration of educational domain rules.

\subsection*{Ablation Analysis and the Role of F-Logic Reasoning}

To systematically evaluate the contribution of each core component in EduRiskX, we conduct an ablation study comparing four model variants:
\begin{enumerate}[label=(\arabic*), leftmargin=*]
    \item \textbf{No Temporal Attention}: The temporal attention module is removed from the neural predictor.
    \item \textbf{No Class-Weighted Loss}: Standard cross-entropy loss replaces the class-weighted loss.
    \item \textbf{EduRiskX (Neural Only)}: The optimized Transformer is used without the F-Logic reasoning module.
    \item \textbf{EduRiskX (Hybrid)}: The complete framework integrating both neural prediction and F-Logic reasoning.
\end{enumerate}

\begin{figure}[htbp]
\centering
\includegraphics[width=0.8\textwidth]{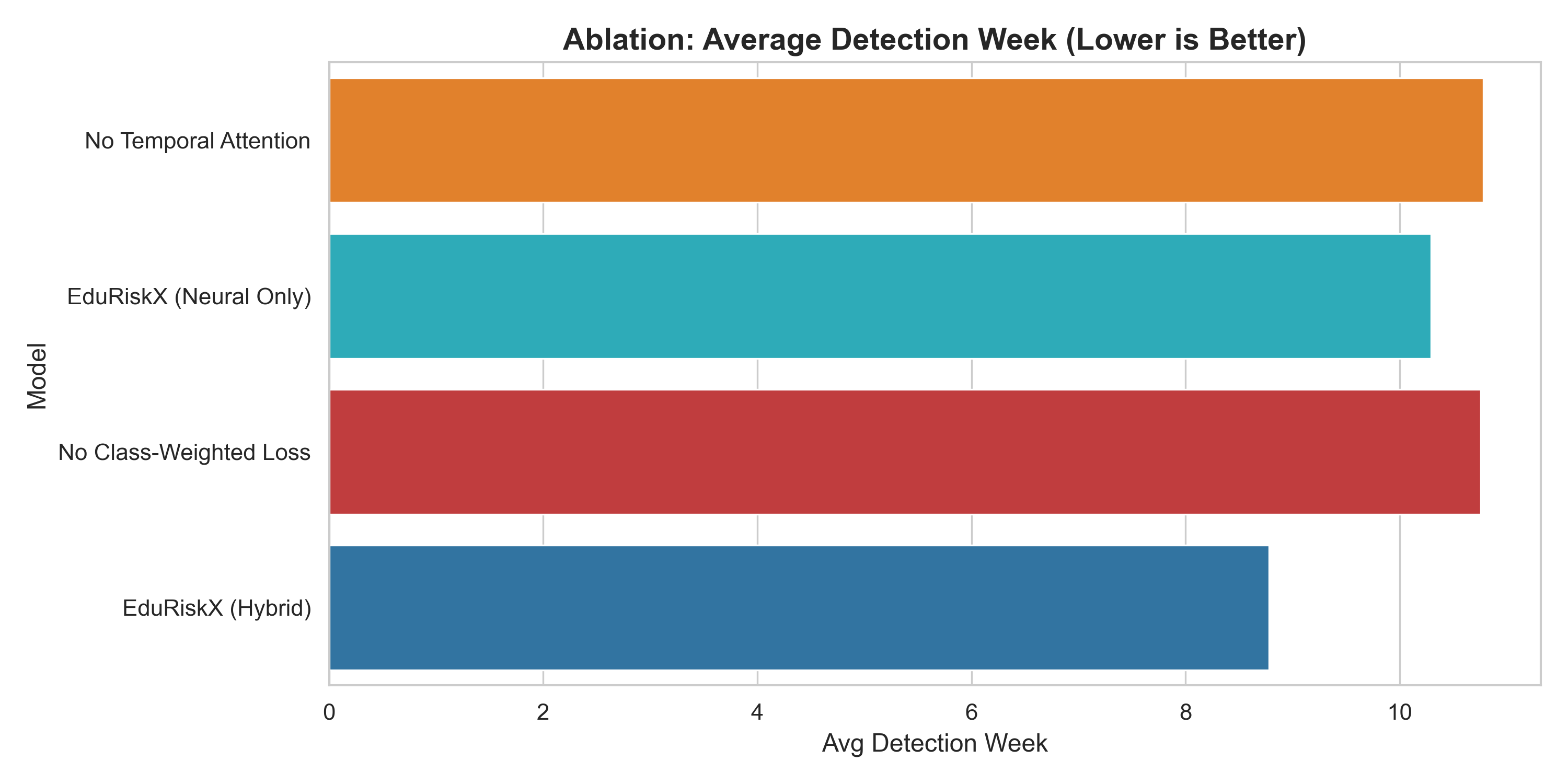}
\caption{\textbf{Average Detection Week Comparison}. EduRiskX (Hybrid) detects at‑risk students 1.5 weeks earlier than the Neural‑Only variant and nearly 2 weeks earlier than other ablation baselines.}
\label{fig:ablation_detection}
\end{figure}

The experimental results reveal the distinct roles of each component:

\begin{enumerate}[label=(\arabic*), leftmargin=*]
    \item \textbf{Temporal Attention}: Removing temporal attention causes a substantial decline in recall, particularly during early weeks (Week~5 recall drops from 0.700 to 0.589). This confirms the module's effectiveness in capturing long-range dependencies from sparse behavioral sequences.
    
    \item \textbf{Class-Weighted Loss}: The omission of class-weighted loss leads to the most pronounced degradation in recall among all variants (Week~5 recall falls to 0.580; Week~38 recall decreases to 0.843). This underscores its critical role in mitigating the severe class imbalance inherent in dropout prediction.
    
    \item \textbf{F-Logic Reasoning}: The Neural‑Only variant (without F‑Logic) exhibits a marked delay in early risk detection. As shown in Figure~\ref{fig:ablation_detection}, the average detection week increases from \textbf{8.78 weeks} (Hybrid) to \textbf{10.29 weeks} (Neural Only)—a delay of approximately 1.5 weeks. This result demonstrates that symbolic reasoning is essential for achieving truly early warnings.
    
    \item \textbf{Overall Performance}: The complete Hybrid model achieves the best balance across all metrics, maintaining consistently high F1‑scores throughout the semester while maximizing early recall.
\end{enumerate}

Figure~\ref{fig:ablation_detection} quantifies the early‑detection advantage: the Hybrid model identifies at‑risk students on average by Week 8.78, significantly earlier than all ablation variants (whose average detection weeks exceed 10 weeks).

\begin{figure}[htbp]
\centering
\includegraphics[width=0.8\textwidth]{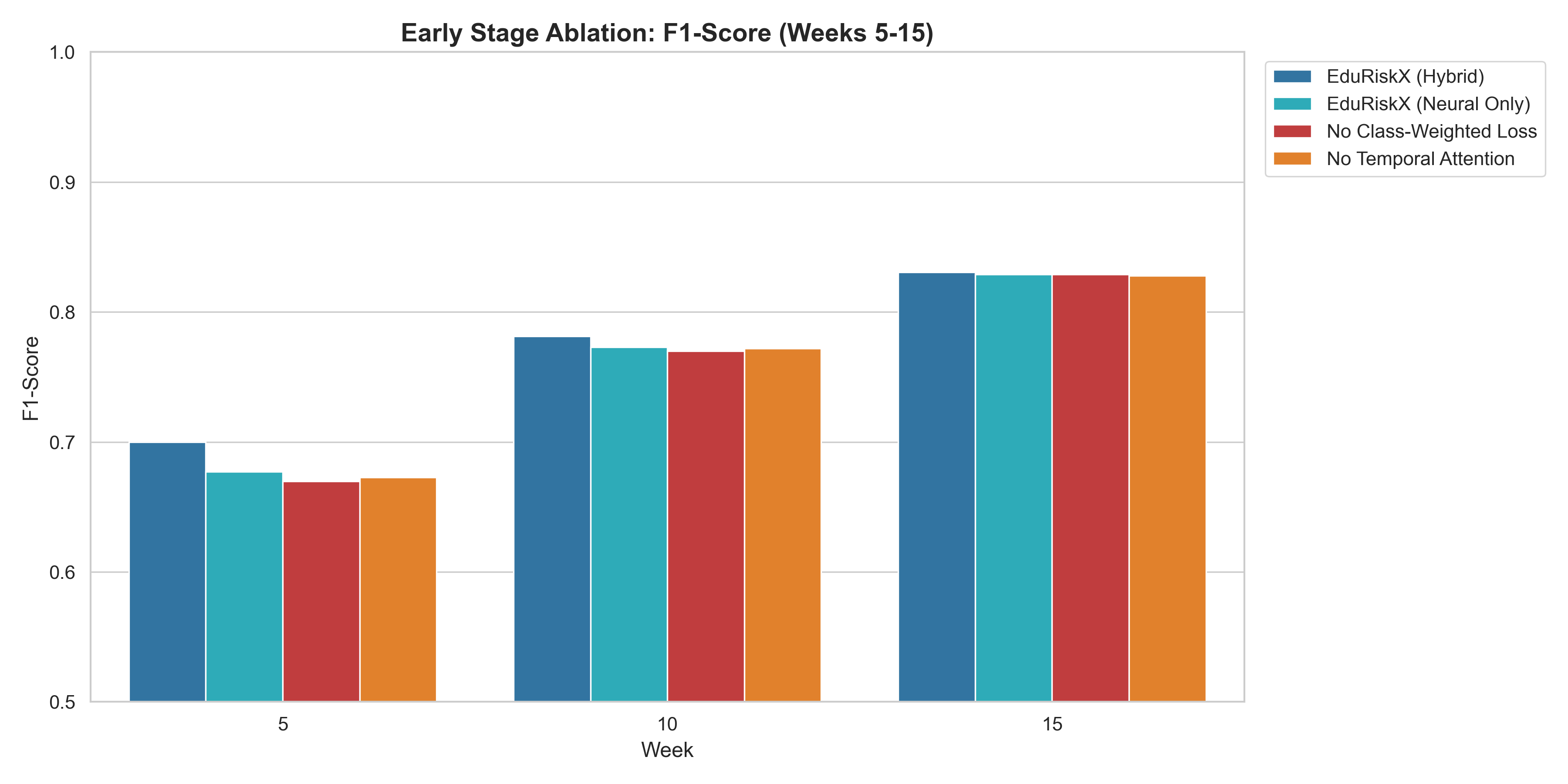}
\caption{\textbf{Early‑Stage F1‑Score Comparison (Weeks 5–15)}. The Hybrid model demonstrates a clear performance edge in the earliest weeks (Week 5 and 10), highlighting its robustness to data sparsity.}
\label{fig:ablation_f1}
\end{figure}

We further examine performance during the critical ``cold start'' phase (Weeks 5–15). Figure~\ref{fig:ablation_f1} presents the F1‑score evolution. At Week 5, the Hybrid model attains an F1‑score of \textbf{0.700}, outperforming the Neural‑Only (0.677) and No‑Temporal‑Attention (0.673) variants. This early advantage indicates that rule‑based reasoning effectively compensates for the limited feature richness of deep learning models in the initial weeks.

\begin{figure}[htbp]
\centering
\includegraphics[width=0.6\textwidth]{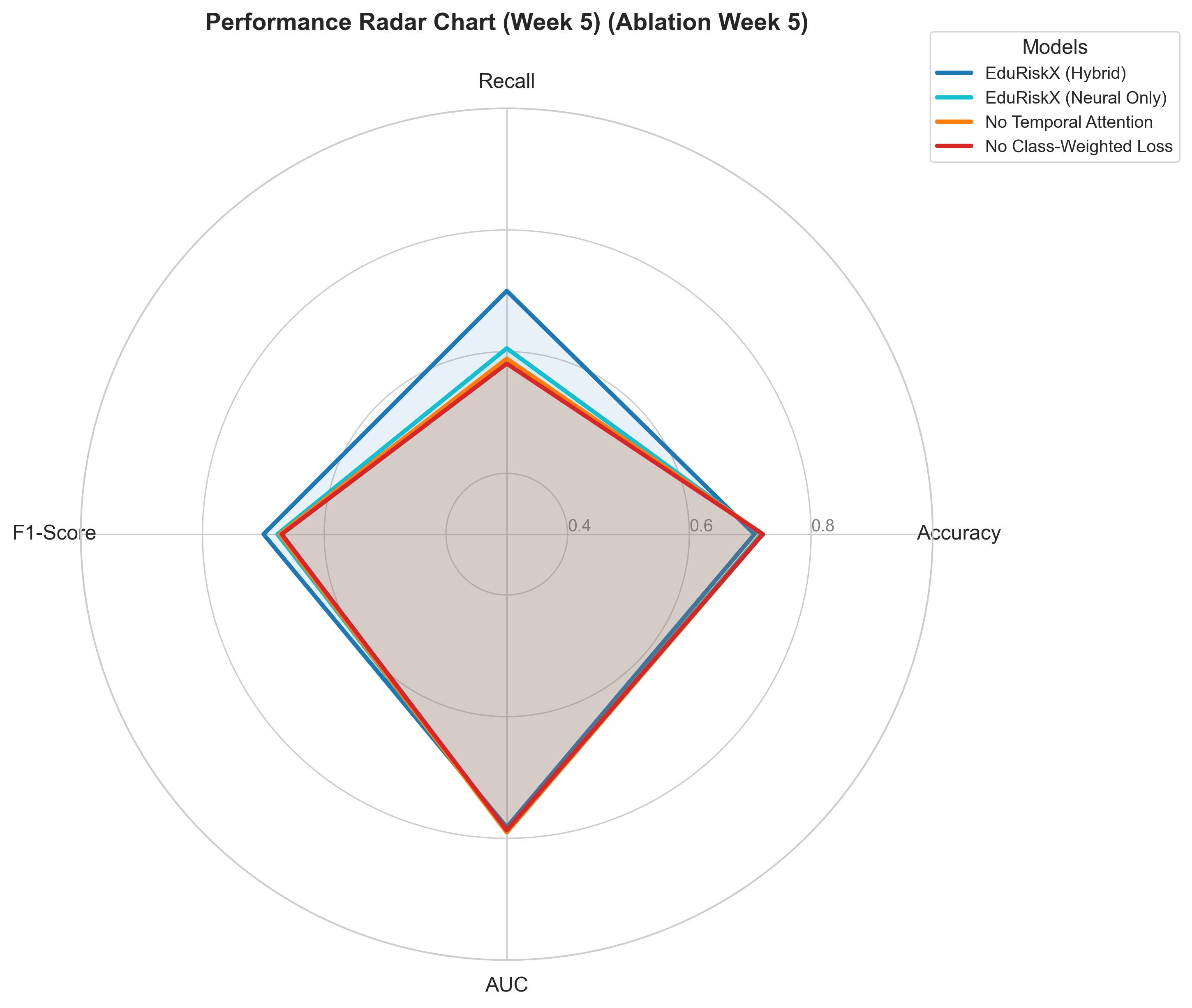}
\caption{\textbf{Week 5 Radar Chart}. EduRiskX (Hybrid) shows comprehensive superiority during the earliest detection phase, particularly in recall and F1‑score.}
\label{fig:ablation_radar}
\end{figure}

To visualize the multidimensional trade‑offs in the early stage, Figure~\ref{fig:ablation_radar} provides a radar chart at Week 5. The Hybrid model (blue area) fully encompasses the other variants, exhibiting superior performance not only in a single metric but simultaneously across accuracy, recall, and AUC.

\begin{figure}[htbp]
\centering
\includegraphics[width=0.8\textwidth]{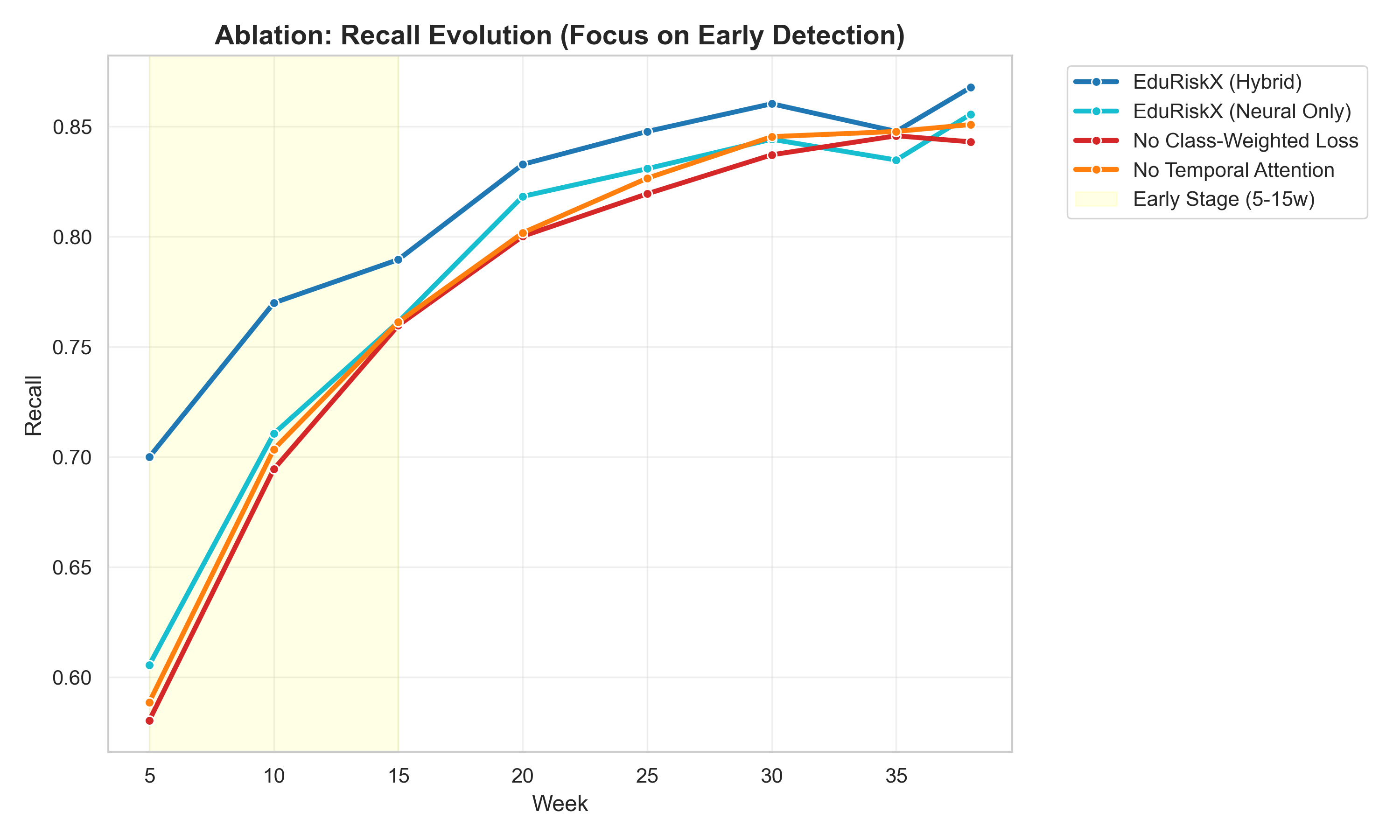}
\caption{\textbf{Recall Evolution (Weeks 5–38)}. The yellow shaded region highlights the early stage, where the Hybrid model maintains a significant recall margin over ablation variants.}
\label{fig:ablation_recall}
\end{figure}

Finally, Figure~\ref{fig:ablation_recall} traces recall evolution over the entire semester. The Hybrid model's curve (blue) consistently remains above those of the ablation variants, especially during the first half of the semester (highlighted in yellow). This sustained high recall is a crucial characteristic for an early warning system, ensuring that the fewest possible at‑risk students are missed during the intervention window.

\subsection*{Training Stability and Early‑Stage Robustness}

We further analyze prediction stability during the early weeks. Pure neural models (Neural Only, No Temporal Attention) exhibit higher variance in predicted risk probabilities during Weeks 5–10, reflecting sensitivity to limited data. In contrast, EduRiskX (Hybrid) produces more stable predictions; the F‑Logic rule‑based reasoning acts as a stabilizer, constraining fluctuations caused by noisy or sparse behavioral signals.

This stability is particularly important in educational settings, where unreliable early warnings can erode trust. By combining data‑driven learning with symbolic constraints, EduRiskX achieves a balance between adaptability and robustness, making it better suited for real‑world early warning deployment.

\subsection*{Statistical Significance Analysis}
\label{subsec:significance}

To evaluate the robustness of model comparisons, we conduct paired t-tests over five independent random student-level splits. For each split, all models are trained and evaluated under identical 80/10/10 train–validation–test partitioning. End-of-semester performance metrics are recorded for each run, and paired comparisons are performed between EduRiskX and each baseline model using split-level scores.

Table~\ref{tab:significance_accuracy} reports the statistical results for end-of-semester accuracy. Compared with the state-of-the-art (SOTA) time-series model PatchTST, EduRiskX achieves a mean accuracy improvement of 2.61\% ($t = 9.51$, $p < 0.001$), indicating statistically significant improvement across random splits. Compared with CNN, the mean improvement is 1.33\% ($t = 3.98$, $p = 0.008$), which is statistically significant at the 0.01 level.
 
For LSTM, the mean accuracy difference is 0.18\% ($p = 0.089$), which does not reach conventional significance thresholds. Similarly, the baseline Transformer shows comparable end-of-semester accuracy performance ($p = 0.951$), suggesting no statistically significant difference.

Although the accuracy difference between EduRiskX and LSTM is not statistically significant, EduRiskX demonstrates statistically significant improvement in F1-score over LSTM ($p = 0.026$), indicating improved balance between precision and recall in risk identification.

Overall, the statistical analysis suggests that the integration of symbolic reasoning with neural prediction contributes to stable and statistically supported improvements over several baseline models, particularly in comparison with strong time-series architectures.

\begin{table}[htbp]
\centering
\caption{Paired t-test results for end-of-semester accuracy over five random splits}
\label{tab:significance_accuracy}
\begin{tabular}{lcccc}
\toprule
Comparison & Mean Diff (\%) & t-statistic & p-value & Significance \\
\midrule
EduRiskX vs PatchTST & +2.61 & 9.51 & $< 0.001$ & *** \\
EduRiskX vs CNN & +1.33 & 3.98 & 0.008 & ** \\
EduRiskX vs LSTM & +0.18 & 1.63 & 0.089 & ns \\
EduRiskX vs Transformer & -0.12 & -2.15 & 0.951 & ns \\
\bottomrule
\end{tabular}
\end{table}

\section*{Case Study}
\label{sec:case_study}

To illustrate how EduRiskX operates in practice and how its F-Logic-based explanations complement neural network predictions, we present representative student-level case studies drawn from the test set of the OULAD dataset.

\subsection*{High-Risk Student with Multi-Rule Evidence}

Student \#27891 represents a classical high-risk example. The Transformer-based neural predictor assigns a risk probability of 0.98 ($\ProbNeural$), indicating high risk. EduRiskX activates the F-Logic reasoning engine and identifies eight triggered rules spanning multiple educational dimensions:
\begin{itemize}[leftmargin=*]
    \item \textbf{Engagement rules}: Continuous decline in weekly activity (STAT019), irregular participation patterns (STAT013);
    \item \textbf{Assessment rules}: Submission delay $> 2$ weeks (TW108), quiz score $< 40$ (TW079);
    \item \textbf{Social integration rules}: No forum participation for 4 consecutive weeks (STAT007).
\end{itemize}

The rule-based risk score ($\ProbRule$) is calculated as 0.75, and the final risk probability using Probabilistic OR is:
\[
\ProbFinal = 0.98 + 0.75 - (0.98 \times 0.75) = 0.995
\]

The framework generates a structured explanation linking these rules to educational theories (low self-efficacy \cite{bandura1997self, locke1997self}, poor academic integration \cite{tinto1975dropout}) and recommends targeted interventions: scaffolded assessment tasks, weekly progress check-ins, and structured peer interaction activities. This case demonstrates that EduRiskX does not merely confirm risk, but contextualizes it within established theoretical constructs, enabling educators to tailor interventions to specific behavioral dimensions.

\subsection*{Correcting Neural Ambiguity with Rule-Based Evidence}

Student \#45122 exemplifies a critical scenario where symbolic reasoning corrects neural network ambiguity. During the first five weeks, this student exhibited contradictory behavioral patterns: stable access to learning resources (suggesting sustained interest) but zero forum participation and delayed first quiz submission. In conventional models, such patterns often lead to misclassification.

Based on the first five weeks of data, the Transformer-based neural predictor yielded a risk probability of $\ProbNeural = 0.42$, below the conventional high-risk threshold (e.g., 0.5). The neural network, influenced by the positive signal of ``stable resource access,'' could not confidently interpret the early risk indicators of ``zero social participation'' and ``submission delay.''

Despite the modest neural prediction, the risk-gating mechanism (with a low $\tau_b^-$) activated rule-based reasoning. Three critical early-window rules were triggered:
\begin{itemize}[leftmargin=*]
    \item \textbf{TW001 (Zero Forum Participation)}: $\text{week} \in [0,4] \land \text{clicks\_forum} = 0$, aligned with \textit{lack of social learning integration} (Student Integration Model) \cite{tinto1975dropout}.
    \item \textbf{TW108 (First Assessment Delay)}: $\text{week} = 4 \land \text{submission\_delay} > 1$, aligned with \textit{deficient self-regulation} (Self-Efficacy) \cite{bandura1997self, locke1997self}.
    \item \textbf{STAT019 (Low Activity Consistency)}: $\text{activity\_consistency} \leq 1.0$, aligned with \textit{fragile behavioral patterns} (Engagement Theory) \cite{fredricks2004school}.
\end{itemize}
These three rules contributed incremental risks of $\Delta p = 0.15 + 0.20 + 0.10 = 0.45$, resulting in $\ProbRule = 0.45$.

The final risk probability was computed using Equation~\eqref{eq:logistic_fusion}:
\[
\ProbFinal = \sigma\big(\alpha \cdot \text{logit}(0.42) + \beta \cdot \text{logit}(0.45) + b\big) = 0.681.
\]
Although the neural prediction was ambiguous (0.42), strong rule-based evidence (0.45) elevated the final risk to a clear ``medium-high risk'' level (0.681).

The generated explanation explicitly stated: \textit{``Although the student demonstrates basic resource access habits, they exhibit complete absence from course community interaction, delayed first assessment submission, and unstable learning rhythm. This aligns with theoretical risk profiles of 'academic isolation' and 'early deficiency in self-regulation.'} Consequently, targeted interventions were adjusted to:
\begin{itemize}[leftmargin=*]
    \item \textbf{Ice-breaking guidance}: Teaching assistants proactively invite the student to join study groups or forum discussions.
    \item \textbf{Time-management support}: Provide personalized assignment planning tools and reminders.
    \item \textbf{Early formative feedback}: Offer detailed, timely feedback on the first assignment.
\end{itemize}

This case vividly demonstrates EduRiskX's ``double safety net'' mechanism. It shows that even when the neural model ``hesitates'' due to sparse data, the pedagogically-grounded rule system can capture critical, theory-informed risk signals, thereby preventing under-reporting of ``quiet but at-risk'' students and enabling truly early warning.

\begin{figure}[htbp]
\centering
\includegraphics[width=0.95\textwidth]{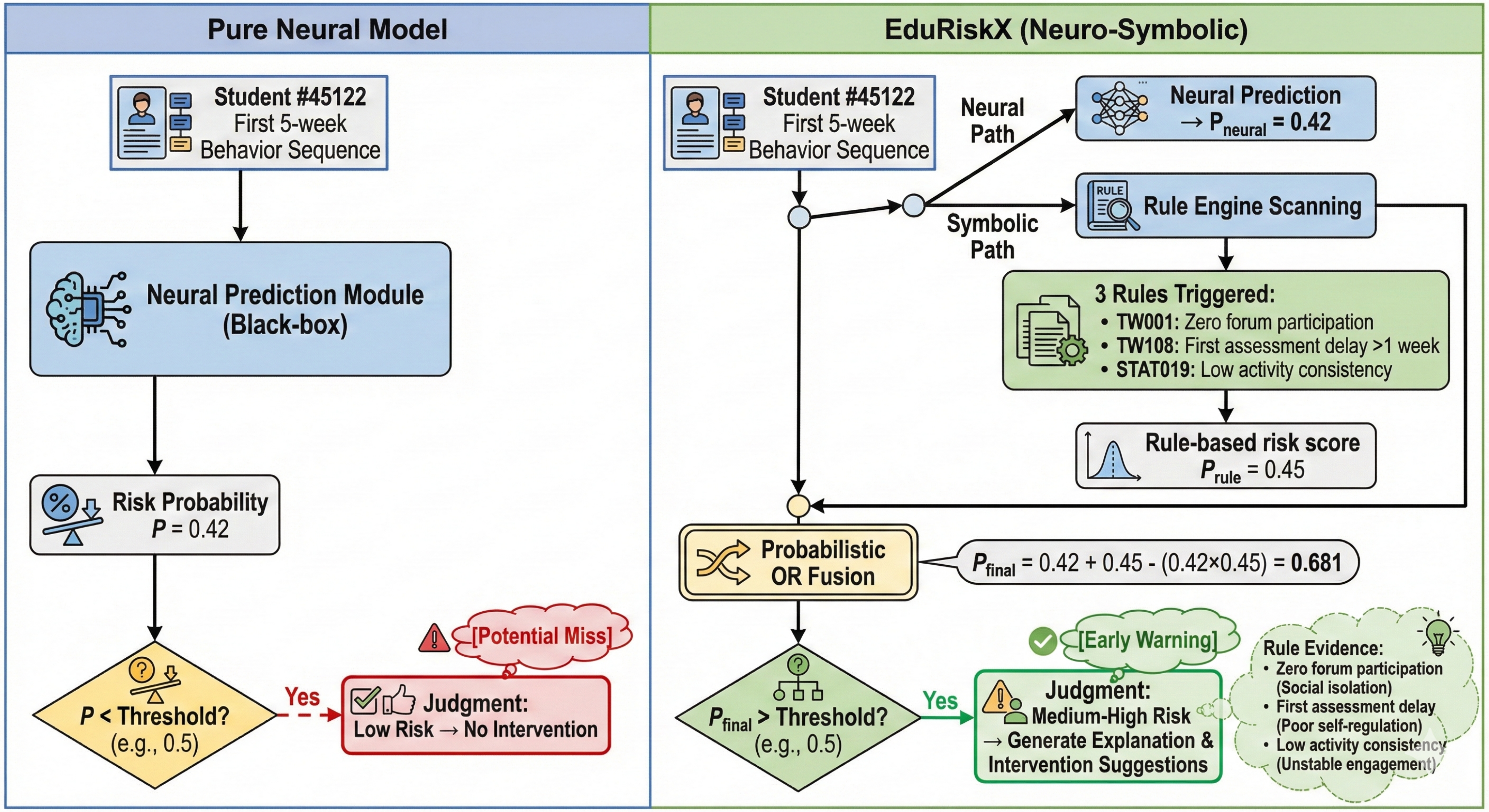}
\caption{Decision flow comparison between EduRiskX and a pure neural model on a borderline case (Student \#45122). The left branch illustrates the pure neural model's decision path, leading to potential under-reporting due to ambiguous early data. The right branch demonstrates how EduRiskX's neuro-symbolic architecture combines neural prediction with rule-based evidence through Probabilistic OR fusion, resulting in correct risk identification and actionable explanations.}
\label{fig:case_study_decision_flow}
\end{figure}

Figure \ref{fig:case_study_decision_flow} illustrates the decision-making contrast between a pure neural model and the EduRiskX neuro-symbolic framework when processing the borderline case of Student \#45122. The diagram employs a side-by-side flowchart layout to visually deconstruct how each system interprets the same early behavioral sequence (first five weeks). The left branch depicts the conventional pipeline of a pure neural predictor (e.g., LSTM or Transformer): the student's activity sequence passes through a black-box neural module, yielding a risk probability of $P = 0.42$, which falls below a typical intervention threshold (e.g., 0.5). Consequently, the model classifies the student as low-risk, leading to no intervention---a potential missed detection highlighted in red.

The right branch elucidates EduRiskX's ``predict-then-explain'' architecture. Here, the same input activates two parallel pathways: a neural prediction path (upper) that outputs $P_{\text{neural}} = 0.42$, and a symbolic reasoning path (lower) where the F-Logic rule engine scans the data and triggers three pedagogically-grounded rules (TW001, TW108, STAT019), aggregating to a rule-based risk score of $P_{\text{rule}} = 0.45$. These two evidence streams converge at the logistic fusion module, which calculates the final risk probability ($P_{\text{final}} = 0.681$) using the hybrid inference formula. Since this value exceeds the threshold, EduRiskX correctly raises a medium-high risk alert and generates an interpretable explanation—citing specific rule violations such as ``zero forum participation'' and ``submission delay''—along with tailored intervention suggestions. The successful early warning is marked in green, directly contrasting with the pure model's under-reporting.

This visual comparison underscores a key contribution of our work: by integrating symbolic reasoning with neural prediction, EduRiskX creates a double safety net that can overcome the ambiguity of data-driven models in early stages, thereby transforming potential misses into actionable, explainable alerts.

\subsection*{Borderline Risk Student and Early Warning Refinement}

Student \#32930 illustrates a borderline case where F-Logic reasoning plays a critical role. The neural predictor assigns a moderate risk probability of 0.65 ($\ProbNeural$), close to the decision threshold. EduRiskX identifies two persistent behavioral patterns:
\begin{itemize}[leftmargin=*]
    \item Continuous decline in content access (STAT019),
    \item Low quiz participation (TW081).
\end{itemize}

The rule-based risk score ($\ProbRule$) is 0.55, and the final risk probability is:
\[
\ProbFinal = 0.65 + 0.55 - (0.65 \times 0.55) = 0.8075
\]

Although only two rules are triggered, their temporal persistence provides sufficient evidence to confirm risk, resulting in earlier detection (Week 8 vs. Week 11 for the neural-only model). The recommended interventions focus on increasing interaction frequency and reinforcing learning routines, reflecting a preventive support strategy rather than intensive remediation.

\subsection*{Low-Risk Student with No Rule Activation}

Student \#26192 exemplifies a low-risk case. The neural predictor assigns a low risk probability of 0.07 ($\ProbNeural$), and no F-Logic rules are triggered ($\ProbRule = 0$). The final risk probability is 0.07, and no intervention is suggested. This behavior demonstrates that EduRiskX does not introduce unnecessary complexity or noise for stable, low-risk students, preserving trust in the system's alerts.

Together, these case studies illustrate that EduRiskX enhances predictive analytics by providing interpretable, theory-aligned explanations that support differentiated and timely educational interventions, particularly in high-risk and borderline scenarios.

\subsection*{Practical Deployment and System Usability}
\label{subsec:practical_deployment}

While the aforementioned case studies illustrate the underlying predictive mechanism of EduRiskX, its core contribution to the field of learning analytics lies in its readiness for real-world deployment as a modern expert system. Traditional pure deep learning models often face a ``black-box'' trust crisis among educators, who are understandably hesitant to initiate interventions based solely on an opaque probability score. EduRiskX overcomes this barrier by generating theory-aligned, human-readable evidence.

From a system deployment perspective, the architecture is highly computationally efficient. The F-Logic reasoning engine operates primarily through logical rule matching rather than dense matrix multiplication. Consequently, the computational cost during inference is negligible, allowing the system to process thousands of student behavioral sequences in real-time or via lightweight weekly batch processing on standard institutional servers. 

In a practical pedagogical setting, the structured JSON outputs generated by the framework (as shown in Figure~\ref{fig:json_output}) are translated into an intuitive visual Dashboard for instructors and academic advisors. Through this user interface, educators do not merely receive a binary ``high-risk'' alert. Instead, they can interact with the system to trace the specific triggered rules (e.g., ``zero forum participation for 4 consecutive weeks''), understand the underlying theoretical deficits (e.g., ``Social Isolation''), and seamlessly review, approve, or adjust the system's automated intervention suggestions. By mimicking the diagnostic reasoning of human educational experts, EduRiskX bridges the gap between complex algorithmic predictions and actionable teaching strategies, significantly increasing its actual deployment value in intelligent educational environments.

\section*{Discussion}
\label{sec:discussion}

This study examines whether integrating a temporal Transformer-based predictor with F-Logic symbolic reasoning can support earlier and more interpretable identification of at-risk students. The results indicate that EduRiskX achieves earlier detection (average 9.32 weeks) compared with strong neural baselines such as PatchTST (15.70 weeks) and iTransformer (13.92 weeks). From a practical perspective, this difference corresponds to a longer potential intervention window within a semester.

\subsection*{Hybrid Modeling and Early-Stage Stability}

One observation from the experiments is that purely data-driven models, including advanced Transformer-based architectures, tend to benefit from increased behavioral observations over time. In early course stages, student activity traces may be sparse or variable, which can affect prediction stability.

By incorporating an F-Logic reasoning module constructed from the training set and grounded in established educational constructs, EduRiskX introduces structured rule-based signals that complement neural predictions. The logistic regression–based fusion mechanism allows neural probabilities and symbolic confidence scores to be combined in a data-driven manner. This design appears to improve prediction consistency in early weeks while maintaining competitive overall performance. The results suggest that incorporating structured domain knowledge may contribute to temporal robustness in early warning tasks.

\subsection*{Interpretability and Educational Use}

In addition to predictive outcomes, EduRiskX produces structured explanations linking risk predictions to specific F-Logic rules and pedagogical constructs. Such explanations provide interpretable evidence that may assist educators in understanding the behavioral basis of a risk alert.

Early warning systems are most useful when they support actionable decisions. Earlier identification, as observed in the experimental results, may provide instructors with additional time to implement supportive measures during formative stages of learning. While the actual effectiveness of interventions depends on institutional context and instructional design, the availability of interpretable early signals may enhance practical usability.

\subsection*{Temporal Evaluation Considerations}

The findings highlight the importance of evaluating predictive models not only by end-of-semester accuracy but also by their temporal detection characteristics. Some baseline models achieve strong final accuracy yet exhibit later stabilization of predictions. EduRiskX demonstrates that incorporating symbolic reasoning and learnable fusion can shift detection earlier in the semester while maintaining comparable overall metrics.

Although the improvement in recall relative to neural-only models is modest in absolute percentage terms, its temporal positioning may have practical implications in educational contexts where intervention timing is critical. These results suggest that early detection week and temporal stability may serve as complementary evaluation criteria in future learning analytics research.

\subsection*{Implications for Learning Analytics Research}

The present study indicates that neuro-symbolic architectures represent a viable direction for educational predictive modeling. In settings where behavioral data are sequential and potentially sparse in early stages, symbolic rule-based components may provide structured guidance that complements neural representation learning.

More broadly, the results suggest that predictive performance and interpretability can be jointly considered through modular hybrid design. Rather than viewing these objectives as strictly competing, carefully designed integration strategies may enable models to address both dimensions in educational early warning systems.

\section*{Conclusion}
\label{sec:conclusion}

This paper presented EduRiskX, a neuro-symbolic framework that integrates a temporal Transformer-based predictor with F-Logic symbolic reasoning for early academic risk prediction in online higher education. The framework explicitly separates neural prediction and symbolic explanation, while combining their outputs through a logistic regression–based fusion mechanism that adaptively learns the contribution of each component.

Experiments conducted on the OULAD dataset using a strict 80/10/10 student-level split demonstrate that EduRiskX achieves competitive end-of-semester performance (accuracy 0.900, F1-score 0.894, recall 0.864) and enables early identification of at-risk students, with an average detection week of 9.32. Compared with state-of-the-art (SOTA) time-series models and widely used deep learning baselines, EduRiskX yields improved recall and earlier detection under identical experimental conditions. Ablation analysis indicates that temporal attention contributes to modeling long-range behavioral dependencies, class-weighted loss mitigates class imbalance, and F-Logic reasoning enhances early-stage recall and interpretability.

Beyond quantitative performance, EduRiskX emphasizes pedagogical transparency and practical deployment value. By functioning as a modern expert system, it resolves the ``black-box'' trust crisis often faced by deep learning models. By grounding symbolic rules in established educational constructs and mimicking the diagnostic logic of human educators, the framework provides structured explanations that connect risk predictions with observable behavioral patterns, supporting interpretable and theory-aligned early intervention. These findings suggest that neuro-symbolic integration with learnable fusion provides a viable approach for developing trustworthy and effective educational early warning systems.

Future work will explore semi-automatic rule refinement, alternative calibration strategies, and validation across additional educational datasets to further assess generalizability.

\section*{Author contributions statement}
Y.F. conceived the study, developed the EduRiskX neuro-symbolic framework, and performed all experimental evaluations on the OULAD dataset. Y.F. was responsible for the construction of the F-Logic rule base and the implementation of the optimized temporal Transformer and logistic fusion mechanism. Y.Z. and R.B. supervised the research and provided pedagogical grounding for the symbolic rules. Y.K. contributed to data preprocessing and temporal feature construction. Y.F. wrote the main manuscript text and prepared all figures and tables. All authors reviewed and approved the final manuscript.

\section*{Funding}
This work was supported by the National Natural Science Foundation of China (NSFC) under Grant No. 62177007.

\section*{Data availability}
The Open University Learning Analytics Dataset (OULAD) analyzed during the current study is publicly available in  UC Irvine Machine Learning Respository. The URL is \url{https://archive.ics.uci.edu/dataset/349/open+university+learning+analytics+dataset}. 

\section*{Code availability}
The source code for the EduRiskX framework is publicly available at Zenodo: \url{https://doi.org/10.5281/zenodo.18637901}.

\section*{Additional information}
\textbf{Competing interests}: The authors declare no competing interests.

\end{document}